\documentclass{article}

\usepackage{microtype}
\usepackage{graphicx}
\usepackage{subcaption}
\usepackage{booktabs} % for professional tables
\usepackage{xcolor} % For colors, useful for emphasis
\usepackage{algorithm} % For algorithm environment
\colorlet{DarkBlue}{blue!70!black}
\colorlet{DarkRed}{red!70!black}
\usepackage[table]{xcolor}
\usepackage[most]{tcolorbox}
\usepackage{lipsum}
\usepackage{multirow}
\usepackage{fancyhdr}       % header
\usepackage{booktabs}
\usepackage{amsmath}
\usepackage{adjustbox}
\usepackage{makecell} % 【重要】需要添加这个包来实现表头换行
\usepackage{hyperref}

\usepackage[preprint]{icml2026}

\usepackage{amsmath}
\usepackage{amssymb}
\usepackage{mathtools}
\usepackage{amsthm}

\usepackage[capitalize,noabbrev]{cleveref}

\theoremstyle{plain}
\newtheorem{theorem}{Theorem}[section]
\newtheorem{proposition}[theorem]{Proposition}

\theoremstyle{definition}

\theoremstyle{remark}

\usepackage{xcolor} % 必加载：文字上色宏包

\definecolor{lightgray}{gray}{0.9} % 定义置灰颜色
\usepackage[textsize=tiny]{todonotes}

\begin{document}

\onecolumn
  % \icmltitle{Accelerating Embodied Policies in World Models}
  \icmltitle{Speedup Patch: Learning a Plug-and-Play Policy to \\Accelerate Embodied Manipulation}



  \icmlsetsymbol{equal}{*}
  \icmlsetsymbol{corres}{$\dagger$}

% \author{Zhichao Wu$^{1,2}$\footnotemark[1],\;
% Junyin Ye$^{1,2}$\footnotemark[1],\;
% Zhilong Zhang$^{1,2}$\footnotemark[1],\;
% Yihao Sun$^{3}$, \; 
% Haoxin Lin$^{2}$, \\
% \textbf{Haoxiang Ren$^{2}$, \; 
% Jiaheng Luo$^{4}$, \;
% Lei Yuan$^{1,2}$, \; 
% Yang Yu$^{1,2}$\footnotemark[2]} \\[1.5ex]
% $^1$National Key Laboratory for Novel Software Technology, Nanjing University, China \\
% $^2$School of Artificial Intelligence, Nanjing University, China \\
% $^3$Mila-Quebec AI Institute \& Université de Montréal, Canada \\
% $^4$Kuang Yaming Honors School, Nanjing University, China \\
% \texttt{\{wuzc,yejy,zhangzl,linhx\}@lamda.nju.edu.cn} \\
% \texttt{\{221294016,241240049\}@smail.nju.edu.cn} \\
% \texttt{yihao.sun@mila.quebec} \\
% \texttt{\{yuanl,yuy\}@nju.edu.cn}
% }
  \begin{icmlauthorlist}
    \icmlauthor{Zhichao Wu}{yyy,comp,equal}
    \icmlauthor{Junyin Ye}{yyy,comp,equal}
    \icmlauthor{Zhilong Zhang}{yyy,comp,equal}
    \icmlauthor{Yihao Sun}{sch}
    \icmlauthor{Haoxin Lin}{yyy,comp}\\
    \icmlauthor{Jiaheng Luo}{yyy,comp}
    \icmlauthor{Haoxiang Ren}{yyy,comp}
    \icmlauthor{Lei Yuan}{yyy,comp}
    \icmlauthor{Yang Yu}{yyy,comp,corres}
  \end{icmlauthorlist}

  \icmlaffiliation{yyy}{National Key Laboratory for Novel Software Technology, Nanjing University, China}
  \icmlaffiliation{comp}{School of Artificial Intelligence, Nanjing University, China}
  \icmlaffiliation{sch}{Mila-Quebec AI Institute \& Université de Montréal, Canada}
  \icmlcorrespondingauthor{Yang Yu}{yuy@nju.edu.cn}

  \icmlkeywords{Machine Learning, ICML}
  \vskip 0.3in



\printAffiliationsAndNotice{}  % no special notice (required even if empty)

\begin{abstract}
While current embodied policies exhibit remarkable manipulation skills, their execution remains unsatisfactorily slow as they inherit the tardy pacing of human demonstrations. Existing acceleration methods typically require policy retraining or costly online interactions, limiting their scalability for large-scale foundation models. In this paper, we propose \textbf{S}peed\textbf{u}p \textbf{P}atch (\textbf{SuP}), a lightweight, policy-agnostic framework that enables \textbf{plug-and-play acceleration} using solely offline data. SuP introduces an external scheduler that adaptively downsamples action chunks provided by embodied policies to eliminate redundancies. Specifically, we formalize the optimization of our scheduler as a Constrained Markov Decision Process (CMDP) aimed at maximizing efficiency without compromising task performance. Since direct success evaluation is infeasible in offline settings, SuP introduces \textbf{World Model based state deviation} as a surrogate metric to enforce safety constraints. By leveraging a learned world model as a virtual evaluator to predict counterfactual trajectories, the scheduler can be optimized via offline reinforcement learning. Empirical results on simulation benchmarks (Libero, Bigym) and real-world tasks validate that SuP achieves an overall $1.8\times$ execution speedup for diverse policies while maintaining their original success rates.

% Despite the success of imitation learning in embodied intelligence, purely mimicking human demonstrations often fails to fully realize the potential of robotic systems. This is most evident in efficiency, where human-centric temporal redundancies bottleneck the intrinsic capabilities of high-performance hardware. In this paper, we propose Speedup Patch (SuP), a model-agnostic framework designed to fulfill the latent potential of robotic systems by enabling plug-and-play execution acceleration using solely offline data. Instead of retraining the backbone, SuP introduces an adaptive external scheduler that strategically downsamples action chunks from pretrained policies to eliminate human-induced motion slack. We formalize this process as a Constrained Markov Decision Process (CMDP) aimed at maximizing system throughput while maintaining task fidelity. To bridge the gap of offline evaluation, SuP leverages a World Model-based state deviation metric as a surrogate for safety, utilizing counterfactual trajectory prediction to guide policy optimization. Empirical results across the LIBERO and Bigym benchmarks, along with real-world robot experiments, demonstrate that SuP unlocks up to $1.4\times$ and $2\times$ execution speedup across diverse architectures without compromising success rates, effectively narrowing the gap between imitation and hardware limits.
\end{abstract}

\section{Introduction}

Recent advances in embodied intelligence have demonstrated the potential of generalizable robotic manipulation. By learning from large-scale demonstration datasets, embodied policies can preform fine-grained manipulation tasks, comprehend natural language instructions, and draw up clear task plans \cite{zhaoaloha, black2410pi0, intelligence2025pi05}. However, despite these capabilities, current embodied policies often suffer from execution inefficiency, completing tasks at a slow pace that hinders real-world deployment \cite{park2024proleptic,guo2025demospeedup}.

% \begin{figure}[!t]
%     \centering
% % \vspace{-2mm}
% \label{fig:intro_method_showcase}
%     \includegraphics[width=1.\linewidth]{figures/intro_method_showcase.pdf}
%     \caption{Our method, Speedup Patch, accelerates embodied manipulation by learning a plug-and-play lightweight policy, while incurring negligible degradation in task success rate during acceleration.}
%     \vspace{-7mm}
% \end{figure}

This execution inefficiency stems primarily from the redundant nature of human teleoperation data, characterized by the slow movement patterns of human demonstrators \cite{guo2025demospeedup}. Consequently, trained policies tend to output excessively dense action sequences, resulting in long task execution times. Existing methods for execution acceleration often come with expensive costs. They necessitate intricate data curation and policy retraining \cite{guo2025demospeedup, kim2025espada}, introduce entirely new action prediction mechanisms \cite{arachchige2025sail}, or rely on expensive online interactions to maintain task success \cite{yuan2025speedtuning, nam2025speedaug}. Such requirements lead to substantial training overhead and limited scalability, especially for large Vision-Language-Action (VLA) models where fine-tuning is computationally expensive. This motivates our core inquiry: \textit{Can we achieve plug-and-play acceleration for diverse embodied policies using solely offline data, without retraining the original policies?}

To bridge the gap, we propose \textbf{S}peed\textbf{u}p \textbf{P}atch (SuP), a lightweight, policy-agnostic framework that learns an external scheduler policy from offline demonstration data alone. The scheduler acts as a ``patch'' that adaptively downsamples the action chunks provided by the embodied policy to eliminate redundancies. We formalize the optimization of our scheduler as a Constrained Markov Decision Process (CMDP) \cite{gu2024review, zhao2023state}, aimed at maximizing execution efficiency without compromising task performance.

However, solving this CMDP in offline setting is non-trivial, as evaluating task success typically necessitates costly online interaction. To avoid this, we propose \textit{World Model based state deviation}—defined as the discrepancy between the end-effector (EEF) trajectories of the original and accelerated actions—as a surrogate safety constraint to bound the acceleration rate. Specifically, we first train a world model to capture the environment dynamics from existing demonstrations. We then leverage this model as a virtual evaluator to generate counterfactual trajectories by simulating various downsampling rates on the offline data. Finally, the scheduler policy is optimized via offline reinforcement learning using these synthesized data, learning to maximize execution speed while strictly bounding state deviation. We empirically validate our method via simulation (Bigym \cite{chernyadev2025bigym} and Libero \cite{liu2023libero}) and real-world experiements, achieving an overall $1.55\times$ and $2.17\times$ execution speedup while maintaining task performance of different embodied policies. Our contributions are summarized as follows:\begin{itemize}
\item We formalize plug-and-play policy acceleration as a CMDP to maximize execution efficiency while preserving policy performance.
\item We solve this CMDP by introducing World Model-based state deviation as a surrogate safety constraint, enabling offline optimization of the external scheduler via counterfactual trajectory evaluation.
\item We empirically validate our method across simulation benchmarks (Libero, Bigym) and real-world robotic platforms, achieving an overall $1.55\times$ and $2.17\times$ speedup while maintaining task performance of different embodied policies.
\end{itemize}
% \begin{itemize}
% \item We propose a plug-and-play, model-agnostic acceleration framework that requires no retraining of the original policy and operates solely on offline data.
% \item We introduce \textit{state deviation} to safely bound the acceleration rate and formulate acceleration as a CMDP, enabling safe, adaptive action downsampling via a lightweight scheduler policy.
% \item We develop an offline model-based RL approach using a learned World Model to train the scheduler without online interaction, achieving state-of-the-art speedups while preserving task performance.
% \end{itemize}

\section{Background}
\subsection{Action Chunking in Embodied Policies}
\label{bg:chunk}
Modern embodied architectures, such as Action Chunking with Transformers \cite{zhao2023learning},  Diffusion Policies \cite{chi2025diffusion} and various Vision-Language Action models \cite{zitkovich2023rt, kim2024openvla, black2410pi0}, have adopted action chunking as a standard paradigm to mitigate compounding errors and improve inference efficiency. Formally, an action chunk is defined as a finite-length sequence of consecutive actions. Let \( n \) denote the chunk length. At each time step $t$, given the current observation $o_t$, a embodied policy $\pi_{base}$ does not predict a single action but rather a chunk of $n$ future actions:
\begin{equation}
    A_t = \left(a_t, a_{t+1}, \dots, a_{t+n-1}\right)
\end{equation}
While effective for fine-grained manipulation, this approach suffers from the high temporal redundancy of human demonstrations. The resulting "step-by-step" execution often leads to sluggish robot behavior, preventing the system from reaching its maximum hardware performance.

\subsection{Action Downsampling for Acceleration}
\label{bg:ds}
% 机器人底层控制系统，可以抽象成为接受command的伺服系统S0，内部通过前馈控制、PID等控制技术实现对目标状态的动态跟踪，以基于位置控制的机器人系统为例，外部输入下一时刻的各个关节的位置qpos_desire，底层控制系统将其转化为高频的电机力矩信号，来使qpos接近qpos_desire，这个过程中tracking error=|qpos-qpos_desire|<epsilon需要delta_system时间，高速移动场景，机器人无法及时跟随指令从而减低精度，惯性带来的动作幅度增大、过冲。高速导致的tracking error增加增大了distribution shift，这就导致了直接加速的方法出现成功率大幅度下滑。如何将tracking error纳入考虑范围，是speedup的主要挑战。
To optimize execution efficiency, action downsampling is employed to reduce the number of physical steps required to complete the task by decimated or interpolating the predicted action chunk. Formally, given a chunk $A_t$ and a downsample rate $k$, action downsampling will produce an accelerated sequence $A_t^k$ of reduced length $l=\lfloor n/k \rfloor$. The specific operation depends on the policy's control mode \cite{lynch2017modern}:
\begin{equation}
    \label{eq:downsample}
    A_t^k =\begin{cases}
        \left(a_{t+k-1}, \dots, a_{t+l k-1}\right) & \text{Abs}\\
        \left(m(a_{t:t+k}), \dots, m(a_{t+l k-k:t+l k})\right) & \text{Delta}
    \end{cases} 
\end{equation} where $m$ is an action-merging function (e.g., summation).
This is semantically reasonable in position-based action spaces because actions directly represent target waypoints; skipping intermediate steps thus maintains the original intent of the trajectory \cite{shi2023waypoint}. However, in real-world robotic control, the actual state reached by low-level controllers rarely coincides perfectly with the target waypoint. Downsampling exacerbates this error by increasing the distance between commanded targets, making it even more difficult to ensure consistency with the original expert trajectory. These deviations in the visited states can accumulate, ultimately leading to task failure.

\section{The Foundation of SuP}

This section formulates plug-and-play speedup as CMDP where a scheduler policy optimizes execution efficiency subject to world model-estimated state deviation constraints.

\subsection{Plug-and-Play Speedup via Scheduler Policy}

\begin{figure}[h]
    \centering
\vspace{-2mm}
    \includegraphics[width=0.7\linewidth]{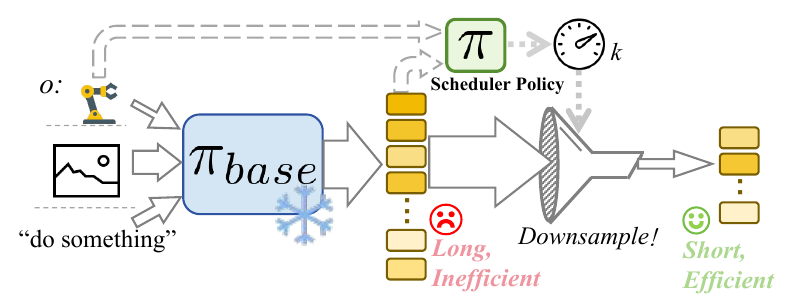}
    \caption{\textbf{Plug-and-Play Speedup via Scheduler Policy.} The scheduler policy $\pi$ predicts a downsampling rate $k$ to downsample the action chunk from the frozen policy into a shorter chunk for acceleration.}
    \label{fig:policy_hierarchy}
    \vspace{-3mm}
\end{figure}

\textbf{Base Policy:} The base policy $\pi_{\text{base}}$ is the policy to be accelerated, which is an visumotor policy capable of predicting action chunk from visual observations and robot states. Specifically, during the inference phase of $\pi_{\text{base}}$, the input consists of the visual observation $I_t$ and robot state $o_t$, and the base policy outputs an action chunk via:
$A_t \sim \pi_{\text{base}}(\cdot | I_t, o_t)$.
Since $\pi_{\text{base}}$ is typically trained on slow demonstration data, the actions predicted by $\pi_{\text{base}}$ are generally inefficient in execution.

\textbf{Scheduler Policy:} We introduce an additional plug-and-play scheduler policy to downsample the action chunk produced by $\pi_{\text{base}}$ for acceleration, as shown in Fig.\ref{fig:policy_hierarchy}. Concretely, the scheduler policy $\pi(\cdot|o_t,A_t)$ is a lightweight policy that predicts a downsample rate $k$ given current state $o_t$ and action chunk $A_t$. The final action executed in the environment is thus the downsampled action chunk $A_t^k$ (Eq. \ref{eq:downsample}). Specifically, when $k=1$, the downsampled action chunk coincides with the original, i.e., $A_t^1 = A_t$. This formulation allows our scheduler $\pi$ to achieve state-dependent execution speedup of $\pi_{\text{base}}$ in a plug-and-play manner.

\subsection{Acceleration via Constrained MDP}
\label{sec:cmdp}

\begin{figure}[!t]
    \centering
    \includegraphics[width=0.95\linewidth]{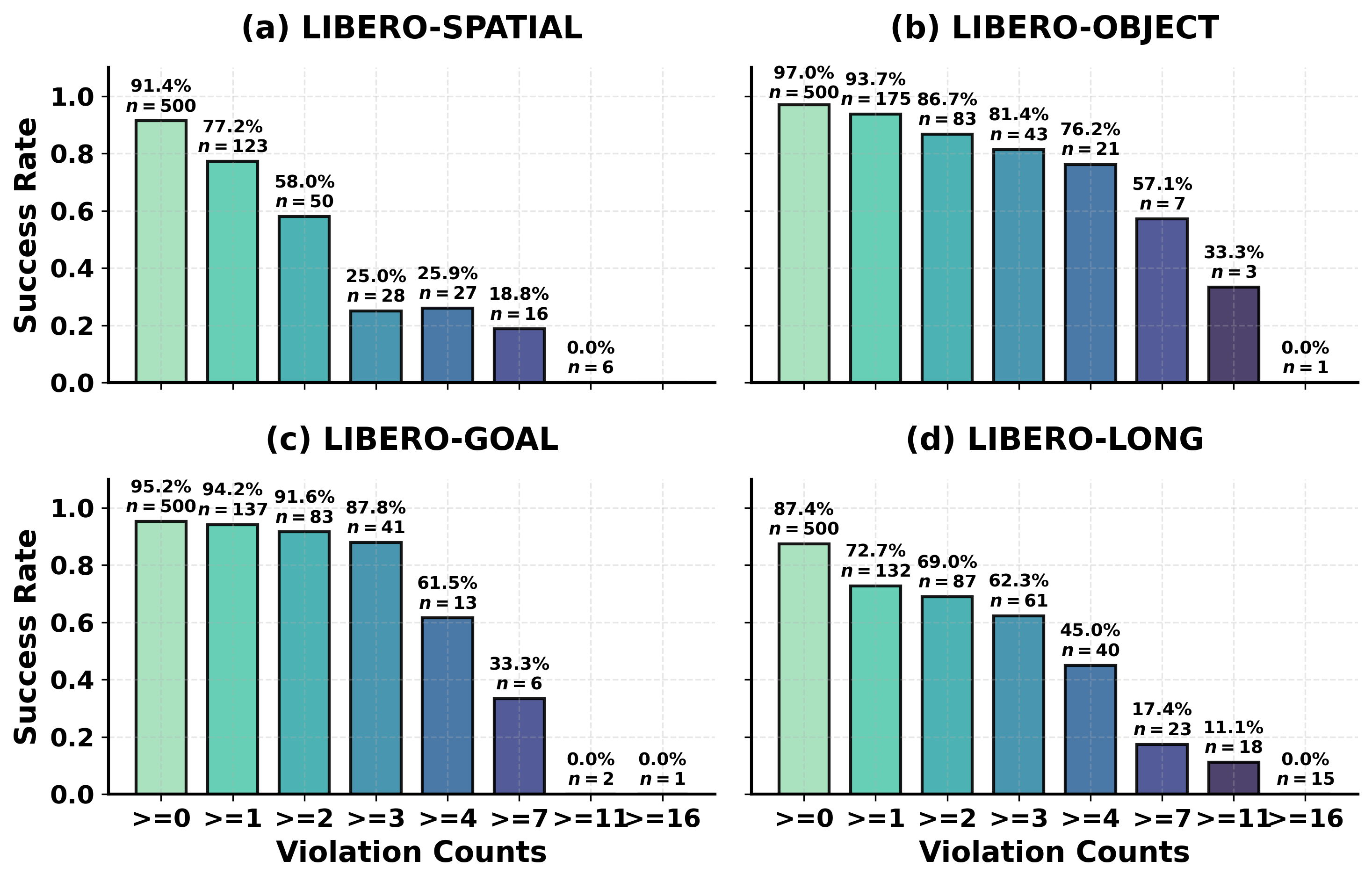}
    \caption{\textbf{Success rate and Violation count.} 
Each subplot (a--d) illustrates the relationship between the cumulative count of violations ($h_\mathcal{E}=1$) and the task success rate across different LIBERO suites. 
The bars represent the conditional success rate for the subset of trajectories containing at least $x$ violations, with the sample size of each subset annotated above the corresponding bar.}
    \label{fig:state_deviation}

\end{figure}
We then formulate speedup learning as a scheduler policy optimization problem within the framework of Constrained Markov Decision Processes (CMDPs), defined by the tuple $(\mathcal{S}, \mathcal{K}, \mathcal{P}, r, c, h, \gamma)$. In this formulation, policy acts as a high-level scheduler that optimizes execution efficiency without compromising task performance. 
$\mathcal{S}$ is the state space augmented to include the current environment observation $o_t$ and the action chunk $A_t$ produced by the base policy $\pi_{\text{base}}$ (i.e., $s_t=(o_t, A_t)$). $\mathcal{K}$ is the action space of our scheduler, defined as a discrete set of downsampling rates $\{k_{\text{min}}, \dots, k_{\text{max}}\}$. $\mathcal{P}$ represents the environment dynamics under the execution of the downsampled action chunk $A_t^k$. To incentivize efficiency, we define the reward function as the acceleration gain: $r(s_t, k_t) = k_t$.

Ideally, the cost function $c$ for acceleration should directly reflect the impact of acceleration on task performance. We define the performance-based cost as:\begin{equation}c_q(s_t, k_t) = Q^{\pi_{\text{base}}}(o_t, A_t^k)-Q^{\pi_{\text{base}}}(o_t, A_t),\end{equation} where $Q^{\pi_{\text{base}}}(o, A)$ represents the expected success rate starting from state $o$ and executing action chunk $A$ from $\pi_{\text{base}}$. The violation function is then $h_q(s_t, k_t) = \mathbb{I}\left[c_{q}(s_t, k_t)<0\right]$. We provide theoretical guarantee as follows:\begin{proposition}\label{prop:performance} Given zero-violation constraint ($h_q(s_t,k_t)=0$) at each state, the scheduler is guaranteed to maintain or improve the success rate of the base policy. See App. \ref{app:proof1} for the proof.\end{proposition}

Therefore, the objective of scheduler is to maximize acceleration gain under a zero-violation constraint:\begin{equation} \begin{aligned} \label{eq:cmdp_final} \max_{\pi} \quad & \mathbb{E}_{\pi} \left[ \sum_{t=0}^{T} \gamma^t r(s_t,k_t) \right] \\ \text{s.t.} \quad & \mathbb{E}_{\pi} \left[ \sum_{t=0}^{T} \gamma^t h_q(s_t, k_t) \right] = 0 \end{aligned} \end{equation}

However, evaluating $c_q$ requires the true value function $Q^{\pi_{\text{base}}}$, which is expensive to estimate online. To enable offline learning, we transition from value-based constraints to state-based constraints using a learned world model.

\subsection{World Model based State Deviation as Cost}

\label{sec:state_deviation}
To evaluate costs offline, we utilize a learned world model $\mathcal{M}_{\theta}$ (Sec. \ref{sec:world_model}) to simulate future trajectories. Our key insight is that the success of $\pi_{\text{base}}$ is tied to its specific motion intent; thus, if the accelerated trajectory remains close to the original one, the task performance is preserved.

Formally, let $\tau_t = \{o_{t+1}, \dots, o_{t+n}\}$ and $\tau^k_t$ be the sequence of states predicted by $\mathcal{M}_\theta$ under the original chunk $A_t$ and the downsampled chunk $A_t^k$. We denote $\hat{\tau}^k_t = \{\hat{o}_{t+1}^k, \dots, \hat{o}_{t+n}^k\}$ as the version of $\tau^k_t$ temporally interpolated to match the length of $\tau_t$. We then define the state deviation $\mathcal{E}$ as the maximum discrepancy between the end-effector (EEF) of these two trajectories: \begin{equation}\label{eq:state_deviation}\mathcal{E}(s_t, k_t) = \max_{i \in [1, n]} d(\text{EEF}(o_{t+i}), \text{EEF}(\hat{o}_{t+i}^k)),\end{equation}where $d(\cdot, \cdot)$ is a distance metric for EEF (App. \ref{app:state_deviation_calc}). 

We empirically validate the reliability of our metric by evaluating the $\pi_{0.5}$ model \cite{intelligence2025pi05} on the LIBERO benchmark with a fixed downsampling rate of $k=2$. As shown in Fig. \ref{fig:state_deviation}, our analysis across four task suites reveals a consistent trend: as the number of violations-defined as $h_\mathcal{E}(s_t, k_t) =\mathbb I\left[\mathcal{E}(s_t,k_t)>\epsilon\right]$—within a trajectory increases, the policy's success rate exhibits an evident decline. This pronounced negative correlation serves as strong evidence that state deviation is a faithful proxy for execution risk, validating its effectiveness as a cost signal.

% We empirically validate this surrogate by evaluating the $\pi_{0.5}$ model \cite{intelligence2025pi05} on the LIBERO benchmark. As shown in Fig. \ref{fig:state_deviation}, our analysis across four task suites reveals a strong negative correlation between the number of violations—defined as $h_\mathcal{E}(s_t, k_t) =\mathbb I\left[\mathcal{E}(s_t,k_t)>\epsilon\right]$—and the final task success rate. This confirms that state deviation $\mathcal{E}$ is a reliable proxy as cost function. Detailed calculations and additional correlation studies are provided in App. [A].

% 跟加速baseline的比较是多维度的： GFLOPS比较，炼丹时长比较，行能比较

\section{Practical Implementation}
\begin{figure*}[t]
    \centering
    \vspace{-2mm}
    \label{fig:method_overview}
    
    \includegraphics[width=0.95\linewidth]{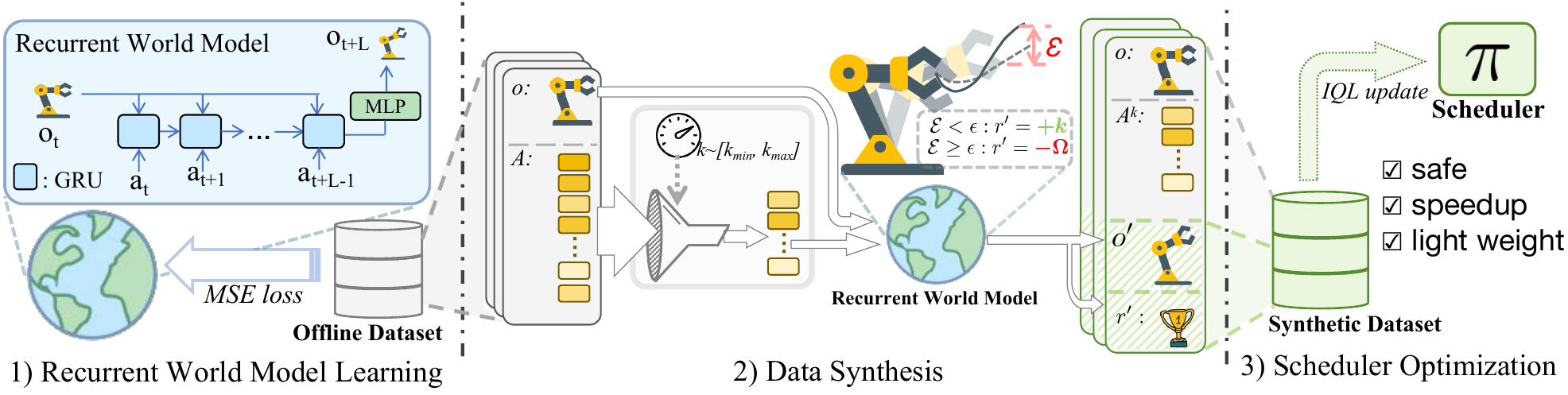}
    \caption{\textbf{The training process of SuP.} Our method is trained purely on offline datasets through three phases: (1) recurrent world model learning; (2) data synthesis; and (3) scheduler optimization via IQL. Through this pipeline, we optimize the scheduler to achieve the maximum possible speedup while preserving comparable performance.}
    \vspace{-3mm}
\end{figure*}

To solve the CMDP for adaptive manipulation acceleration, we propose Speedup Patch (SuP), a dual-stage framework. Our objective is to learn an optimal scheduler policy $\pi_\phi$ that maximizes execution speed while maintaining task fidelity from an offline demonstration dataset $\mathcal{D}$. We first train a Recurrent World Model (RWM) on $\mathcal{D}$ (Sec. \ref{sec:world_model}), allowing us to evaluate potential state deviations and constraint violations. Using the RWM as a data evaluator and generator, we synthesize a CMDP dataset. We then optimize the final scheduler via Implicit Q-Learning (Sec. \ref{sec:scheduler}).

\subsection{Recurrent World Model}
\label{sec:world_model}
To estimate the state deviation $\mathcal{E}$ and the violation signal $h$ in an offline setting, we develop a Recurrent World Model (RWM), denoted as $\mathcal{M}_\theta$. To ensure the framework remains lightweight, $\mathcal{M}_\theta$ is designed to predict the robot's state $o$ directly, thereby bypassing the high-dimensional reconstruction of visual observations $I$.

Our RWM architecture is built upon ADM \cite{lin2024any}, which is specifically capable of handling variable-length action sequences while mitigating the compounding errors typical in multi-step rollouts. A pivotal feature of this design is that predicted states $\hat{o}$ are never fed back as inputs for subsequent time steps. Instead, the model evolves its hidden state exclusively through the action sequence. 

The RWM is optimized via variable-length sequence supervision. During training, we sample action sequences $A_t$ of variable length $L \in [1, L_{max}]$ from the offline dataset $\mathcal{D}$, where the range of $L$ is chosen to encompass the temporal scales generated by different downsampling factors $k$. The parameters $\theta$ are updated to minimize the multi-step Mean Squared Error:
\begin{equation}\label{eq:wm_loss}\mathcal{L}(\theta) = \mathbb{E}_{(o_t, A_t, o_{t+1:t+L}) \sim \mathcal{D}} \left[ \sum_{i=1}^{L} \|\hat{o}_{t+i} - o_{t+i}\|^2_2 \right].\end{equation}

\subsection{Scheduler Optimization}
\label{sec:scheduler}
With the world model $\mathcal{M}_\theta$, we now describe the learning process for the scheduler policy $\pi_{\phi}$. To solve the CMDP, we first transform the constrained problem into an unconstrained MDP via a penalty-augmented reward:
\begin{equation}
\label{eq:reward_new}
r'(s,k) = \begin{cases} k, & h_\mathcal{E}(s,k)=0 \\ -\Omega, & h_\mathcal{E}(s,k)=1 \end{cases}
\end{equation}
where $\Omega$ is a sufficiently large penalty to guarantee zero-violation safety (Prop. \ref{prop:penalty}). To facilitate offline training, we construct a synthetic dataset $\mathcal{D}'$ by re-labeling the original demonstrations. Specifically, for each transition in $\mathcal{D}$, we use $\mathcal{M}_\theta$ to evaluate the violation signal $h$ across all potential downsampling factors $k$, generating a rich set of counterfactual transitions $(o, A^k, r', o')$.

\begin{proposition}
\label{prop:penalty}
Let $K_{\max}$ be the maximum possible speedup rate and $\gamma \in [0, 1)$ be the discount factor. If the penalty $\Omega$ satisfies the condition:
\begin{equation}
\Omega > \frac{\gamma K_{\max}}{1-\gamma},
\end{equation}
then the optimal policy $\pi^*$ maximizing the cumulative reward of $r'(s, k)$ satisfies the constraint $h(s, \pi^*(s)) = 0$ for all reachable states $s$. See App. \ref{app:proof2} for the proof.
\end{proposition}

In a standard RL setting, computing temporal difference (TD) errors requires the next action $A'$, which is unavailable since we do not know how expert will behave with $s'$. We resolve this by transforming the policy input: instead of directly processing the raw tuple $(s, k)$, the scheduler maps the action chunk $A$ and skip-length $k$ into a downsampled representation $A^k$. By conditioning the policy on $(o, A^k)$, we can treat the resulting $s'$ directly as the subsequent state in a Markovian transition, effectively bypassing the need for future action chunks during value estimation.

Finally, we employ Implicit Q-Learning (IQL) \cite{kostrikovoffline} to optimize the scheduler on $\mathcal{D}'$. The value function $V_\psi$ and Q-function $Q_\phi$ are learned via expectile regression:\begin{equation}
\label{eq:iql}
\begin{aligned}
    L_Q(\phi)&=\mathbb{E}_{(o,A^k,r',o')\sim D'}[(r'+\gamma V_\psi (o')-Q_\phi(o,A^k))^2],\\
    L_V(\psi)&=\mathbb{E}_{(o,A^k)\sim D'}[L^\alpha_2( V_\psi (o)-Q_\phi(o,A^k))],
\end{aligned}
\end{equation}where $L^\alpha_2(x)=|\alpha-\mathbb{I}(x<0)|x^2$ is the expectile loss. During inference, the optimal skip-length is determined by $\pi_\phi(o,A)=\arg\max_k Q_\phi(o,A^k)$.

\begin{algorithm}[t]
   \caption{Training Procedure of SuP}
   \label{alg:sup_training}
\begin{algorithmic}[1]
   \STATE {\bfseries Input:} Offline demonstration dataset $\mathcal{D}$, minimum and maximum downsampling rate $k_{\min},k_{\max}$ penalty $\Omega$, deviation threshold $\epsilon$.
   \STATE {\bfseries Output:} Scheduler policy $\pi_{\phi}$.
   
   \STATE \COMMENT{\textbf{Phase 1: Recurrent World Model Learning}}
   \STATE Initialize world model $\mathcal{M}_\theta$.
   \WHILE{not converged}
      \STATE Sample batch of data $(o_t, A_t, o_{t+1:t+L})$ from $\mathcal{D}$.
      \STATE Update $\mathcal{M}_\theta$ with Eq. \ref{eq:wm_loss}.
   \ENDWHILE

   \STATE \COMMENT{\textbf{Phase 2: Data Synthesis}}
   \STATE Initialize synthetic dataset $\mathcal{D}' \leftarrow \emptyset$.
   \FOR{each transition $(o_t, A_t)$ in $\mathcal{D}$}
      \FOR{$k = k_{\min}$ {\bfseries to} $k_{\max}$}
         \STATE Construct downsampled action chunk $A^k_t$.
         \STATE Predict next states $\hat{o}'_{t+1:t+L} \leftarrow \mathcal{M}_\theta(o, A^k)$.
         \STATE Estimate deviation $\mathcal{E}$ with Eq. \ref{eq:state_deviation} and violation signal $h_\mathcal{E} \leftarrow \mathbb{I}(\mathcal{E} > \epsilon)$.
         \STATE Compute reward $r'_t$ with Eq. \ref{eq:reward_new}.
         \STATE Store transition $(o_t, A^k_t, r'_t, \hat{o}'_{t+L})$ into $\mathcal{D}'$.
      \ENDFOR
   \ENDFOR

   \STATE \COMMENT{\textbf{Phase 3: Scheduler Optimization via IQL}}
   \STATE Initialize IQL networks $V_\psi, Q_\phi$.
   \WHILE{not converged}
      \STATE Sample batch $(o, A^k, r', o')$ from $\mathcal{D}'$.
      \STATE Update $V_\psi,Q_\phi$ with Eq. \ref{eq:iql}.

   \ENDWHILE
   \STATE \textbf{Return} Scheduler $\pi_\phi(o, A) = \arg\max_k Q_\phi(o, A^k)$.
\end{algorithmic}
\end{algorithm}
\vspace{-5mm}

\section{Experiments}
%In this section, we通过实验系统评估我们的算法SuP。我们需要回答：
% 1）
% 2）
\definecolor{HighLightColor}{rgb}{1.0, 0.92, 0.8}
\definecolor{lightgray}{gray}{0.9} % 定义置灰颜色
\newcommand{\cg}[1]{{\cellcolor{HighLightColor}#1}}
\newcommand{\bg}[1]{{\cellcolor{lightgray}#1}}
In this section, we conduct extensive experiments to evaluate the effectiveness of the proposed SuP framework. Specifically, we aim to investigate: (1) whether SuP can achieve significant execution speedup while preserving task success rates (Sec. \ref{sec:sim_exp}); (2) the versatility of SuP across different embodied architectures of $\pi_{\text{base}}$ (Sec. \ref{sec:sim_exp}); (3) SuP's empirical performance and reliability in real-world robotic experiments (Sec. \ref{sec:real_exp}); (4) the mechanism by which SuP dynamically selects appropriate downsampling ratios across diverse task scenarios (Sec. \ref{sec:case_exp}); and (5) the individual contributions of RWM, IQL, and different deviation threshold $\epsilon$ settings to the overall system performance (Sec. \ref{sec:ablation}).
    
\subsection{Simulation Task Experiments}
\label{sec:sim_exp}

\begin{figure}[h]
    \centering
% \vspace{-2mm}
\label{fig:simulation_task}
    \includegraphics[width=0.95\linewidth]{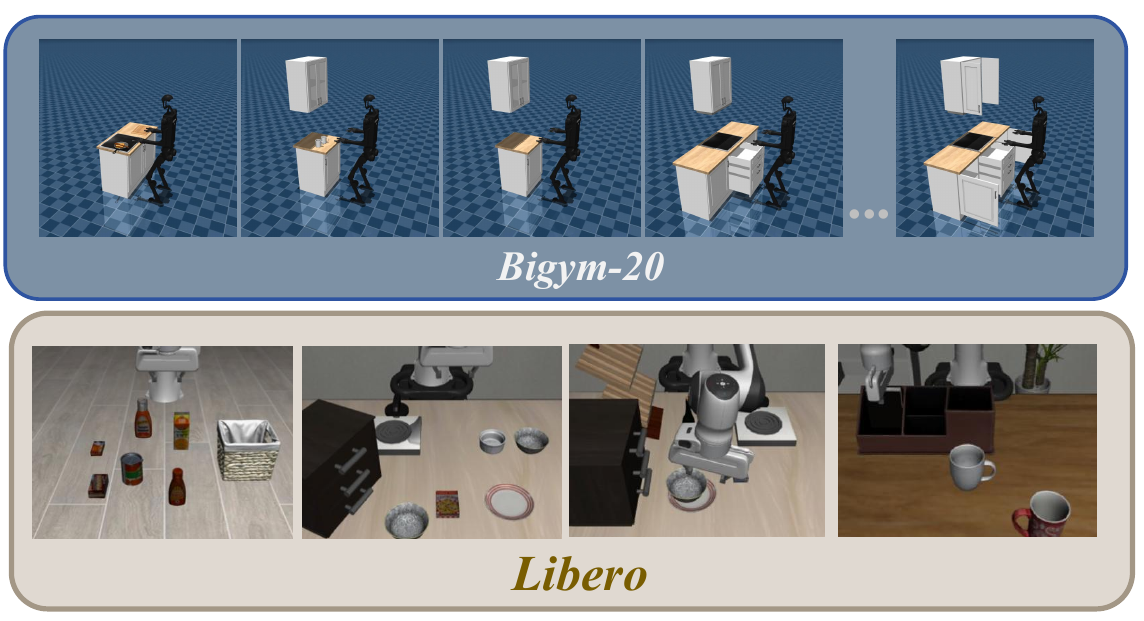}
    \caption{\textbf{Simulation Tasks.} We systematically evaluate SuP across 20 tasks from Bigym and 4 task suites (40 tasks in total) from Libero.}
    \vspace{-5mm}
\end{figure}

\textbf{Compared methods.}~For the baseline methods considered in our comparative experiments, we select the following approaches:
\textbf{Vanilla Downsample(-ds*)}, which applies a fixed downsampling rate to all action chunks and 
\textbf{DemoSpeedup}, which speedup expert demonstration data with entropy estimation to \textbf{retrain} $\pi_{\text{base}}$.

\textbf{Task Setup.}~For simulation tasks, We validate the SuP algorithm with two benchmarks: \textbf{Bigym} \cite{chernyadev2025bigym}, a humanoid robot with kitchen/household manipulation that requires precise control and scene comprehension; \textbf{Libero} \cite{liu2023libero}, a robotic arm grasping benchmark for VLA models, covering 4 task suite, where policies need strong instruction-following abilities. For $\pi_{\text{base}}$, we evaluate our framework across diverse architectures: ACT \cite{zhao2023learning} and DP \cite{chi2025diffusion} trained on task-specific expert demonstrations for Bigym; pre-trained $\pi_{0.5}$ \cite{intelligence2025pi05} and VLA-Adapter \cite{wang2025vla} (weights can be directly downloads from Internet) for Libero. Regarding the scheduler training, we train task-specific SuP schedulers for each Bigym environment, while for Libero, a single scheduler is trained for each task suite.

\begin{table*}[t!]
    \centering
    
    \caption{SuP speedup results on Bigym Tasks compared with baselines. Each cell in the table reports two metrics: the success rate followed by the average steps to completion, where only successful trajectories are counted. Higher success rates and lower step counts indicate better performance. Cells highlighted in \colorbox{HighLightColor}{orange} denote the best success rate for each task, while those in \colorbox{lightgray}{gray} indicate a performance \textbf{drop exceeding 5\%} compared to the best success rate. 4 task results are shown, see \textbf{all task results in App.} 
    \ref{sec:whole_bigym}.}
    \label{tab:bigym_exp}
    
    \small
    \resizebox{\linewidth}{!}{
    \begin{tabular}{clcccc c c c} 
        \toprule
        \multicolumn{2}{c}{\textbf{Method}} & 
        {\small\textbf{\makecell{Sandwich\\Remove}}} & 
        {\small\textbf{\makecell{Take\\Cups}}} & 
        {\small\textbf{\makecell{Put\\Cups}}} & 
        {\small\textbf{\makecell{Drawers\\Close All}}} & 
        {\footnotesize\makecell{(15 more tasks)\\$\dots$}} & 
        {\small\textbf{\makecell{Cupboards\\Close All}}} & 
        \textbf{Average} \\ 
        \midrule
        \multirow{4}{*}{ACT} 
          & -base                & 0.45, 340.5 & 0.15, 288.3 & 0.28, 320.3 & 1.0, 100.0 & $\dots$ & 1.0, 449.8 & 0.66, 1.00$\times$ \\
          & -ds2                 & 0.48, 186.3 & 0.13, 178.0 & 0.36, 175.9 & 1.0, 52.0  & $\dots$ & 1.0, 234.0 & \bg{0.61, 1.65$\times$} \\
          & +\textit{DemoSpeedup}& 0.56, 171.5 & \bg{0.10, 183.9} & 0.33, 169.3 & 1.0, 54.0  & $\dots$ & \cg{1.0, \textbf{202.1}} & \bg{0.61, 2.21$\times$} \\
          & \textbf{+\textit{SuP(Ours)}}  & \cg{\textbf{0.64}, 155.9} & \cg{\textbf{0.20}, 176.6} & \cg{\textbf{0.38}, 156.0} & \cg{1.0, \textbf{40.0}}  & $\dots$ & 1.0, 212.1 & \cg{\textbf{0.67}, \textbf{2.01}$\times$} \\
        \midrule
        \multirow{4}{*}{DP}  
          & -base                & 0.40, 376.8 & 0.07, 284.6 & \cg{\textbf{0.28}, 307.7} & 0.66, 118.9& $\dots$ & 0.90, 544.0 & \cg{0.51, 1.00$\times$} \\
          & -ds2                 & 0.40, 209.6 & 0.12, 218.0 & \bg{0.22, 181.5} & \bg{0.54, 64.9} & $\dots$ & \bg{0.75, 270.3} & \bg{0.40, 1.42$\times$} \\
          & +\textit{DemoSpeedup}& \bg{0.35, 199.3} & \cg{\textbf{0.21}, 239.5} & 0.25, 143.9 & \bg{0.37, 49.2} & $\dots$ & \bg{0.60, 231.0} & \bg{0.46, 1.99$\times$} \\
          & \textbf{+\textit{SuP(Ours)}}  & \cg{\textbf{0.42}, 179.9} & 0.19, 203.0 & 0.27, 200.8 & \cg{0.66, \textbf{65.2}} & $\dots$ & \cg{\textbf{0.91}, 146.2} & \cg{\textbf{0.51}, 1.48$\times$} \\
        \bottomrule
    \end{tabular}
    }
\end{table*}
\begin{table*}[t!]
\centering
% 标题
\caption{SuP speedup results on Libero compared with baselines. Each cell in the table reports two metrics: the success rate followed by the average steps to completion, where only successful trajectories are counted. Higher success rates and lower step counts indicate better performance. Cells highlighted in \colorbox{HighLightColor}{orange} denote the best success rate for each task, while those in \colorbox{lightgray}{gray} indicate a performance \textbf{drop exceeding 1\%} compared to the best success rate.}
\label{tab:libero_exp}
\small
\resizebox{0.9\linewidth}{!}{
\begin{tabular}{clccccc}
    \toprule
    \multicolumn{2}{c}{\textbf{Method}} & \textbf{Spatial} & \textbf{Long} & \textbf{Goal} & \textbf{Object} & \textbf{Average} \\
    \midrule
    \multirow{4}{*}{$\pi_{0.5}$} & -base               & \cg{\textbf{0.988}, 105.3} & 0.924, 267.9 & 0.980, 113.1 & 0.982, 138.1 & 0.969, 1.00$\times$ \\
                            & -ds2                & \bg{0.914, 67.9} & \bg{0.874, 153.4} & \bg{0.952, 67.6} & \bg{0.970, 75.0} & \bg{0.928, 1.72$\times$} \\
                            & +\textit{DemoSpeedup} & \bg{0.964, 88.1} & 0.932, 221.4 & \bg{0.968, 88.7} & 0.988, 114.1 & 0.963, 1.22$\times$ \\
                            & \textbf{+\textit{SuP(Ours)}} & \bg{0.972, 70.4} & \cg{\textbf{0.940}, 215.2} & \cg{\textbf{0.986}, 93.4} & \cg{\textbf{0.994}, 83.0} & \cg{\textbf{0.973}, \textbf{1.35}$\times$}\\
    \midrule
    \multirow{4}{*}{\makecell{VLA-\\Adapter}} & -base                & \cg{\textbf{0.922}, 99.6} & \cg{\textbf{0.936}, 255.1} & \cg{\textbf{0.970}, 107.0} & 0.942, 136.2 & \textbf{0.942}, 1.00$\times$ \\
                         & -ds2                 & \bg{0.802, 57.1} & \bg{0.834, 147.6} & \bg{0.930, 57.1} & \bg{0.882, 76.3} & \bg{0.862, 1.77$\times$} \\
                         & +\textit{DemoSpeedup} & - & - & - & - & - \\
                         & \textbf{+\textit{SuP(Ours)}}  & 0.912, 77.3 & 0.934, 204.6 & \bg{0.956, 74.2} & \cg{\textbf{0.944}, 91.4} & \cg{0.937, \textbf{1.34}$\times$}\\
    \bottomrule
\end{tabular}
}
\end{table*}

\textbf{Metrics.} To evaluate performance, we report the success rate and the average episode length of successful rollouts as a measure of efficiency. We conduct 100 evaluation trials for each task in BiGym, and 500 trials per task suite in Libero. More details of simulation can be found in App. \ref{app:sim_exp_detail}.

\textbf{Speedup Performance.}~The main experimental results on Bigym are presented in Tab. \ref{tab:bigym_exp}, and those on Libero are summarized in Tab. \ref{tab:libero_exp}. Across both challenging benchmarks, SuP demonstrates a superior capability to accelerate inference while maintaining, and often enhancing performance. Unlike baselines such as standard downsampling (-ds2) and DemoSpeedup, which frequently suffer from performance degradation—evidenced by the gray cells indicating a noticeable performance drop—SuP consistently maintains the original performance $\pi_{\text{base}}$. On Bigym, SuP achieves substantial average speedups (e.g., $2.01\times$ for ACT) while maintaining its success rate. Similarly, on Libero, SuP yields a $1.35\times$ speedup for $\pi_{0.5}$ with a peak average success rate of 0.973, effectively decoupling inference speed from performance loss and proving its robustness in numerous simulation tasks.

\textbf{Universality across Architectures.}~SuP exhibits strong generalizability across diverse policy backbones, ranging from ACT and DP to VLAs. For the ACT architecture, SuP not only doubles the inference speed but also improves the average success rate compared to the base policy. Crucially, on DP—which is sensitive to temporal modifications—SuP successfully mitigates the severe performance collapse observed in other acceleration methods (where -ds2 drops success to 0.4), recovering the success rate to 0.51 with a $1.48\times$ speedup. This consistent efficacy extends to VLA architectures while Demospeedup faces \textbf{compatibility issues}, which is inapplicable to VLA-Adapter due to the architecture's lack of support for entropy estimation. In contrast, SuP outperforms naive downsampling strategies on both VLA-Adapter and $\pi_{0.5}$, delivering stable acceleration without compromising decision-making precision.

\textbf{Computational Efficiency.}~As shown in Tab. \ref{tab:compute}, SuP achieves high computational efficiency in both training and inference time. Unlike DemoSpeedup, which incurs high computational costs by training on the massive parameters of VLA models and requiring frequent base policy queries, SuP is exceptionally lightweight with only 5.12M trainable parameters. Crucially, our training strategy completely decouples policy learning from base policy inference; instead of querying $\pi_{\text{base}}$, we optimize our light-weight world model and scheduler using only offline data. This design drastically reduces training time, and the scheduler’s inference overhead (1 ms) is negligible compared with the 50 ms inference latency of the $\pi_{0.5}$.

\begin{table}[h]
  \centering  % 表格居中
  \caption{Computational Efficiency of SuP in Libero.}  % 表格标题
  \label{tab:compute}
  \small
  \begin{tabular}{lccc}
    \toprule
    \textbf{Method} & \textbf{\makecell{Training\\Params}} & \textbf{\makecell{Training\\Time}} & \textbf{\makecell{Inference\\Overhead}} \\
    \midrule
    DemoSpeedup       & 4B     & 20h  &  -  \\
    SuP (Ours)         & 5.12M  & 2h  &  1ms (2\%)  \\
    \bottomrule
  \end{tabular}
\end{table}

% 【直接加速的不可行+wm的准确性+state deviation】
\subsection{Real-world Task Experiments}
\label{sec:real_exp}

\begin{figure*}[h!]
    \centering
    \vspace{-2mm}
    \includegraphics[width=0.98\linewidth]{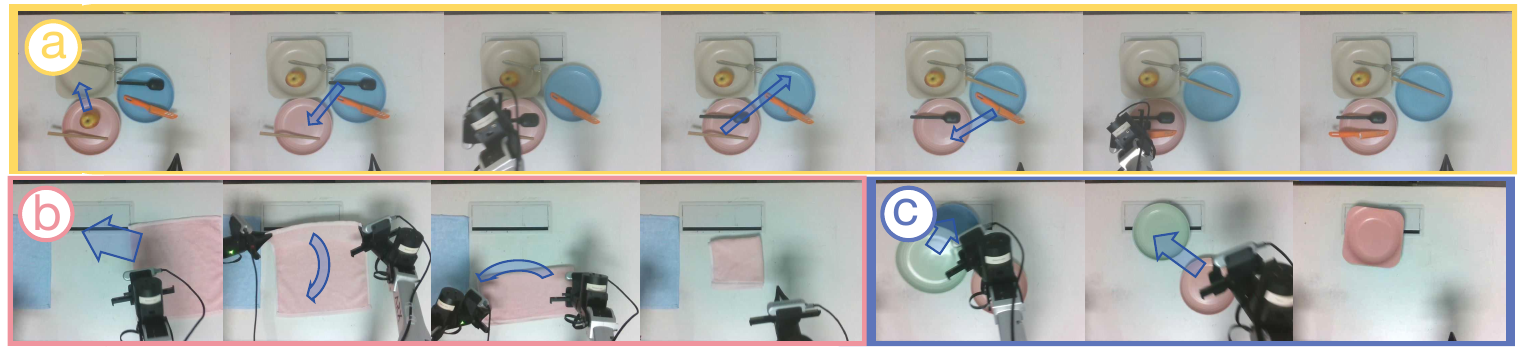}
    \caption{Real-world Tasks Illustration. We illustrate the procedure of 3 real-world tasks: (a) \textit{Arange Table} (b) \textit{Fold Towel} (c) \textit{Stack Plates}.}
    \label{fig:real_task}
    \vspace{-3mm}
\end{figure*}

\begin{table*}[h!]
\centering
% 标题
\caption{SuP speedup results on Real-world Tasks compared with baselines.}
\label{tab:real_exp}
\small
\begin{tabular}{clcccc}
    \toprule
    \multicolumn{2}{c}{\textbf{Method}} & \textbf{Arrange Table} & \textbf{Fold Towel} & \textbf{Stack Plates}  & \textbf{Average} \\
    \midrule
    \multirow{4}{*}{$\pi_{0.5}$} & -base  & 11/30, 537.8 & 15/30, 519.5 & 27/30, 221.5  & 0.589, 1.00$\times$ \\
                            & +\textit{ds2} & 10/30, 291.9 & 14/30, 326.3 & 27/30, 177.1  & 0.567, 1.61$\times$\\
                            & +\textit{ds3} & \bg{3/30, 247.2} & \bg{8/30, 187.7} & \bg{21/30, 148.4}  & \bg{0.356, 2.19$\times$}\\
                            & +\textit{DemoSpeedup} & 12/30, 267.4 & 14/30, 223.2 & \cg{\textbf{28/30}, 124.4}  & 0.600, 2.07$\times$\\
                            & +\textit{SuP(Ours)} & \cg{\textbf{13/30}, 258.5} & \cg{\textbf{16/30}, 192.5} & 26/30, 138.5  & \cg{\textbf{0.611}, \textbf{2.17}}$\times$\\
    \bottomrule
\end{tabular}
% 156.1   91.0 128.1 115.5
\end{table*}

To evaluate the practical efficacy of SuP in physical environments, we deployed SuP on a dual-arm robotic platform—similar to the Aloha setup \cite{zhao2023learning}—focusing on manipulation tasks with multiple steps that require a balance between execution speed and operational success. Our evaluation suite consists of three tasks (Fig. \ref{fig:real_task}): \textit{Arrange Table}, \textit{Fold Towel} and \textit{Stack Plates}. Among these, \textit{Arrange Table} and \textit{Stack Plates} are conducted as single-arm setting, while \textit{Fold Towel} serves as a bimanual task involving deformable object manipulation. Detailed descriptions of the experimental hardware and task setup are provided in App. \ref{app:real_exp}. 

As summarized in Tab. \ref{tab:real_exp}, the results demonstrate that SuP consistently maintains high success rates while achieving significant temporal speedups across all tasks. Specifically, SuP achieves an average speedup of $2.17\times$ over the base policy $\pi_{0.5}$, outperforming the $2.07\times$ of Demospeedup while simultaneously maintaining the original success rate. This performance stands in sharp contrast to naive acceleration strategies; as highlighted in gray, the aggressive ds3 strategy leads to a catastrophic collapse in manipulation capability, dropping the success rate to 0.356. Notably, in the challenging bimanual \textit{Fold Towel} task—which requires precise coordination for deformable objects—SuP attains the highest success rate and the lowest step count, validating its robustness in improving real-world task efficiency.

\subsection{Case Study}
\label{sec:case_exp}
\begin{figure*}[h!]
    \centering
    \vspace{-2mm}
    \includegraphics[width=0.95\linewidth]{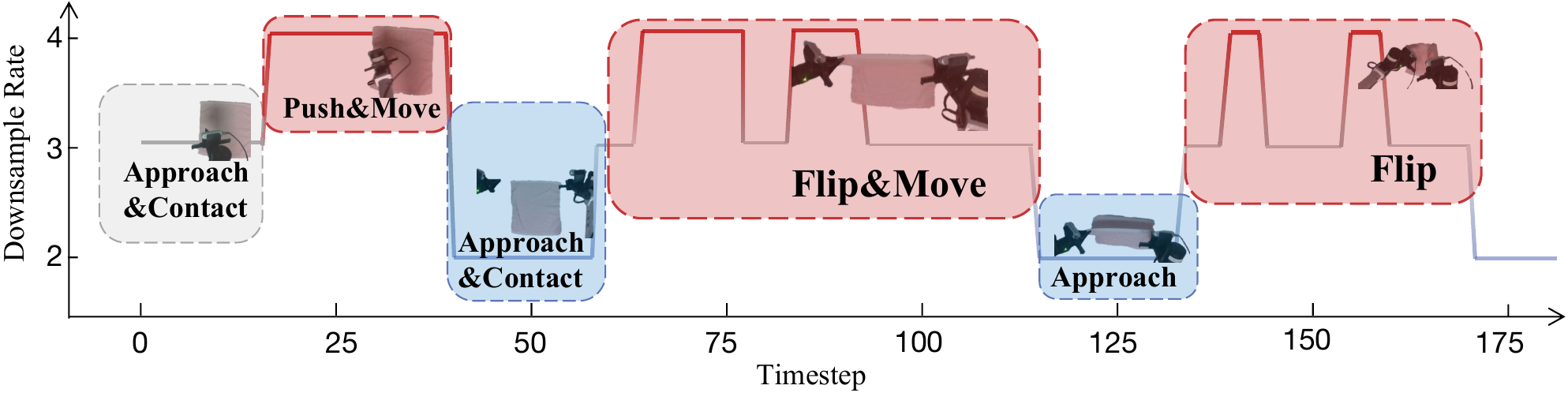}
    \caption{Case study. We visualize the adaptive downsampling strategy during a \textit{Fold Towel} task. The plot tracks the predicted downsampling rate over the episode timesteps. Shaded regions annotate distinct task phases.}
    \label{fig:case_study_real}
    \vspace{-3mm}
\end{figure*}

To analyze how SuP dynamically selects the downsampling rate, we visualize the selected rates during the \textit{Fold Towel} task (Fig. \ref{fig:case_study_real}). In the plot, the blue-shaded regions correspond to phases where the model strictly predicts a low downsampling rate ($k=2$), while red-shaded regions highlight periods of accelerated execution ($k=4$). We observe that this behavior is highly interpretable: the model maintains the low rate during precision-critical phases such as ``Approach \& Contact''. Conversely, during gross motion phases like Push \& Move'' or ``Flip'', the model increases the rate to exploit temporal redundancy. This demonstrates that SuP effectively distinguishes between key decision points and translational phases, accelerating execution without sacrificing control where it matters most.

\subsection{Ablation Study}
\label{sec:ablation}

\begin{figure}[h]
    \centering
    \includegraphics[width=0.99\linewidth]{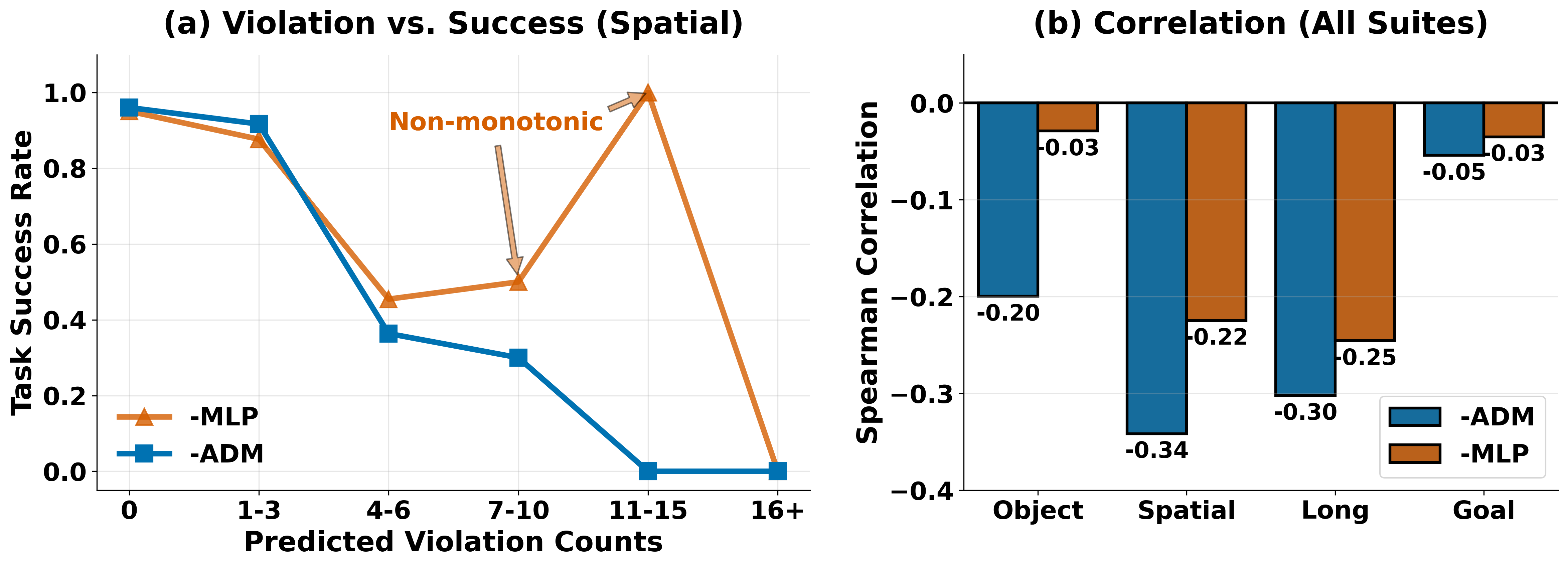}
    \caption{Comparison between ADM-based and MLP-based world model. (a) Visualization of task success rates against predicted violation counts in the Spatial suite. (b) Spearman correlation scores across four LIBERO suites.}
    \label{fig:ablate_adm}
\end{figure}
\textbf{Impact of ADM-based world model.}~To validate the effectiveness of our design, we compare our ADM-based world model against a standard MLP baseline on the Libero benchmark. As illustrated in Fig.~\ref{fig:ablate_adm}, MLP model exhibits inconsistent, non-monotonic behavior in the Spatial suite, failing to correctly associate high violation counts with task failure. In contrast, ADM maintains a monotonic decrease in success rates as violations increase. Furthermore, quantitative analysis across all suites confirms that ADM achieves consistently stronger negative Spearman correlations \cite{spearman1961proof}, demonstrating its superior capability in capturing the inverse relationship between safety violations and task success.

\textbf{Impact of IQL.} We compare SuP with the MPC baseline, which greedily select the highest downsample rate without violation $h_{\epsilon}$. As shown in Tab. \ref{tab:ablation}, the MPC exhibits inferior performance in both success rate and task efficiency. This is because MPC's greedy selection only considers immediate constraints, leading to cumulative errors that compromise long-term stability.

\textbf{Sensitivity on $\epsilon$.}  We evaluate the performance under different deviation thresholds. As shown in Tab. \ref{tab:ablation}, a small threshold ($\epsilon=0.01$) results in conservative behavior and sub-optimal efficiency. Conversely, a large threshold ($\epsilon=0.02$) prioritizes speed but allows excessive deviations, leading to a decline in success rates. Our framework achieves the most robust performance at $\epsilon=0.015$, demonstrating that a moderate threshold effectively triggers RL intervention at the right moment to maintain both stability and optimality across varying tasks.
\begin{table}[h]
\centering
\caption{Ablation study on SuP. We compare the performance impact of the IQL module and different deviation thresholds $\epsilon$.}
\label{tab:ablation}
\small
\begin{tabular}{lcccc}
\toprule
\textbf{Ablation} & \textbf{\makecell{Sandwich\\Remove}} & \textbf{\makecell{Put\\Cup}} & \textbf{Long}\\
\midrule
$\pi_{base}$ & 0.45, 340.5 & 0.28, 320.3 & 0.924, 267.9 \\
\midrule
SuP-0.01 & 0.59, 187.5 & 0.35, 159.8 & \cg{\textbf{0.940}, 215.2} \\
SuP-0.015  &  \cg{\textbf{0.64}, 155.9} & \cg{\textbf{0.38}, 156.0} & 0.922, 191.8 \\
SuP-0.02  & 0.46, 157.6 & \bg{0.21, 143.3} & \bg{0.904, 168.4} \\
\midrule
MPC-0.01 & 0.54, 181.4 & 0.32, 154.8 & 0.936, 225.2  \\
MPC-0.015  &  0.59, 173.3 & 0.3, 146.1 & 0.924, 205.6 \\
MPC-0.02  & \bg{0.35, 162.1} & \bg{0.19, 148.4} & 0.922, 173.7 \\
\bottomrule
\end{tabular}
\vspace{-3mm}
\end{table}

\section{Related Works}
% 在具身智能领域，学习一个或多个操作技能，可以通过对专家数据集进行模仿实现。这里涉及的模仿学习方法，主要是behavior cloning,即对离线数据集中动作标签进行回归或者离散化后的分类拟合。由于动作分布通常具备multi-modal的性质，具身智能领域的模仿学习通常利用强大的生成式的模型结构或方法，如transformer、diffusion model、flow matching等.近年，随着大语言模型的兴起，也有很多模型在VLM上构建出VLA模型，希望利用VLM内部通用的语言、视觉理解能力，从而提高机器人的操作能力。然而，由于数据集中通常是慢速轨迹，导致学到的策略一般速度较慢，限制了具身智能在现实中的进一步落地。

\textbf{Imitation Learning in Embodied AI.}~~
% Learning manipulation skills from human demonstrations has become a dominant paradigm in embodied AI \cite{ravichandar2020recent}, primarily realized through imitation learning which treats policy learning as a supervised regression or classification problem on expert demonstrations \cite{zare2024survey}. To address the multi-modality of human demonstration data, modern IL frameworks increasingly adopt generative backbones. Prominent approaches range from Transformer-based conditional VAEs \cite{zhao2023learning} to diffusion models \cite{chi2025diffusion} and flow matching techniques \cite{lipmanflow} for high-fidelity action chunking. Concurrently, the success of Large Language Models has catalyzed the emergence of Vision-Language-Action (VLA) models \cite{zitkovich2023rt, kim2024openvla, black2410pi0, intelligence2025pi05}, which fine-tune pre-trained VLMs to inject generalized semantic and visual understanding directly into low-level robotic control. Despite their success in success rates and generalization, these methods universally suffer from a critical limitation: they faithfully mimic the temporal characteristics of the training data, thereby failing to fully exploit the execution speed potential inherent to the robotic hardware.
Imitation learning has established itself as a dominant paradigm in embodied AI\cite{zare2024survey, ravichandar2020recent}, evolving from standard regression to sophisticated generative backbones \cite{chi2025diffusion, lipmanflow} and Vision-Language-Action (VLA) models \cite{zitkovich2023rt, kim2024openvla}. Despite their impressive generalization, these methods share a critical limitation: by faithfully mimicking the temporal pacing of human demonstrations, they fail to exploit the full execution speed potential inherent to robotic hardware. Modern embodied AI relies heavily on generative imitation learning \cite{chi2025diffusion} and Vision-Language-Action (VLA) models \cite{zitkovich2023rt, kim2024openvla, black2410pi0, intelligence2025pi05} to handle multi-modal data and semantic understanding. Despite their success in success rates and generalization, these methods universally suffer from a critical limitation: they faithfully mimic the temporal characteristics of the training data, thereby failing to fully exploit the execution speed potential inherent to the robotic hardware.

% \textbf{Fast Policy Execution.}~~Fast execution of robot has been a long-standing goal of robotics. To mitigate the execution inefficiency of standard IL policies, recent research has explored acceleration strategies from both data-centric and model-centric perspectives. Data-centric approaches such as DemoSpeedup \cite{guo2025demospeedup} and ESPADA \cite{kim2025espada} retrain policies on downsampled or filtered datasets, but they suffer from policy retraining, which is computationally expensive for VLAs. Alternatively, model-centric methods introduce specialized temporal modules \cite{arachchige2025sail} or adaptive skipping strategies \cite{yuan2025speedtuning, nam2025speedaug}. However, these typically require architectural modifications or expensive online interactions that pose safety risks. In contrast, our SuP enables plug-and-play acceleration purely from offline data, eliminating the need for retraining or online exploration.

\textbf{Fast Policy Execution.}~Achieving fast policy execution is a longstanding objective in robotics, essential for deployment in dynamic real-world environments \cite{pham2019critically, kiyokawa2022challenges}. Recent work handles the inference latency induced by large VLA models via asynchronous inference \cite{black2025real, black2025training, tang2025vlash}, model compression \cite{yang2025efficientvla, gao2025compressor, wu2025device} or other tricks \cite{ma2025running}. While these methods accelerate individual decision cycles, they overlook the \textit{temporal redundancy} inherent in control horizons. Consequently, recent research has explored reducing the total execution steps (trajectory shortening). These approaches generally fall into data-centric strategies that retrain policies on downsampled demonstrations \cite{guo2025demospeedup, kim2025espada}, or model-centric methods introducing adaptive skipping modules \cite{arachchige2025sail, yuan2025speedtuning}. However, such methods typically mandate computationally expensive retraining or risky online exploration. In contrast, SuP enables plug-and-play acceleration purely from offline data, eliminating these overheads.

% 你是一个具身领域专家，在投稿ICML， 你需要根据中文粗略大意，给出简洁、严谨的英文，最后以latex格式给出，注意公式语法：
\section{Conclusion}

We present Speedup Patch (SuP), a framework that enables plug-and-play execution acceleration for embodied policies using solely offline data. SuP employs a World Model to simulate potential outcomes of accelerated actions, allowing the CMDP-based scheduler to strategically skip redundant time steps without compromising task success rate. Our experiments on simulation and real-world tasks demonstrate that SuP consistently achieves significant speedups across diverse architectures without sacrificing success rates. Future work can move beyond action chunk downsampling-based speedup and design a more flexible, non-integer speedup mechanism to achieve finer-grained speedup.

% 未来工作SuP可以尝试将图像引入world model和scheduler的输入中，测试更多跨任务的泛化能力。
\section*{Impact Statement}
This paper presents work whose goal is to advance the field of Machine Learning. There are many potential societal consequences of our work, none of which we feel must be specifically highlighted here.


\bibliography{example_paper}

@article{guo2025demospeedup,
  title={DemoSpeedup: Accelerating Visuomotor Policies via Entropy-Guided Demonstration Acceleration},
  author={Guo, Lingxiao and Xue, Zhengrong and Xu, Zijing and Xu, Huazhe},
  journal={arXiv preprint arXiv:2506.05064},
  year={2025}
}

@article{kim2025espada,
  title={ESPADA: Execution Speedup via Semantics Aware Demonstration Data Downsampling for Imitation Learning},
  author={Kim, Byungju and Pahk, Jinu and Lee, Chungwoo and Kim, Jaejoon and Lee, Jangha and Kim, Theo Taeyeong and Shim, Kyuhwan and Lee, Jun Ki and Zhang, Byoung-Tak},
  journal={arXiv preprint arXiv:2512.07371},
  year={2025}
}

@inproceedings{arachchige2025sail,
  title={SAIL: Faster-than-Demonstration Execution of Imitation Learning Policies},
  author={Arachchige, Nadun Ranawaka and Chen, Zhenyang and Jung, Wonsuhk and Shin, Woo Chul and Bansal, Rohan and He, Yu Hang and Lin, Yingyan Celine and Joffe, Benjamin and Kousik, Shreyas and Xu, Danfei},
  booktitle={ICRA 2025 Workshop: Beyond Pick and Place},
  year={2025}
}

@inproceedings{yuan2025speedtuning,
  title={Speedtuning: Speeding Up Policy Execution with Lightweight Reinforcement Learning},
  author={Yuan, David D and Zhao, Tony Z and Burns, Kaylee and Finn, Chelsea},
  booktitle={2025 IEEE International Conference on Robotics and Automation (ICRA)},
  pages={1184--1192},
  year={2025},
  organization={IEEE}
}

@article{park2024proleptic,
  title={Proleptic Temporal Ensemble for Improving the Speed of Robot Tasks Generated by Imitation Learning},
  author={Park, Hyeonjun and Lim, Daegyu and Kim, Seungyeon and Park, Sumin},
  journal={arXiv preprint arXiv:2410.16981},
  year={2024}
}

@article{nam2025speedaug,
  title={SpeedAug: Policy Acceleration via Tempo-Enriched Policy and RL Fine-Tuning},
  author={Nam, Taewook and Hwang, Sung Ju},
  journal={arXiv preprint arXiv:2512.00062},
  year={2025}
}

@article{black2410pi0,
  title={$\pi$0: A vision-language-action flow model for general robot control.},
  author={Kevin Black and Noah Brown and Danny Driess and Adnan Esmail and Michael Equi and Chelsea Finn and Niccolo Fusai and Lachy Groom and Karol Hausman and Brian Ichter and Szymon Jakubczak and Tim Jones and Liyiming Ke and Sergey Levine and Adrian Li-Bell and Mohith Mothukuri and Suraj Nair and Karl Pertsch and Lucy Xiaoyang Shi and James Tanner and Quan Vuong and Anna Walling and Haohuan Wang and Ury Zhilinsky},
  journal={arXiv preprint arXiv:2410.24164},
  year={2024}
}

@article{intelligence2025pi05,
      title={$\pi_{0.5}$: a Vision-Language-Action Model with Open-World Generalization}, 
      author={Kevin Black and Noah Brown and James Darpinian and Karan Dhabalia and Danny Driess and Adnan Esmail and Michael Equi and Chelsea Finn and Niccolo Fusai and Manuel Y. Galliker and Dibya Ghosh and Lachy Groom and Karol Hausman and Brian Ichter and Szymon Jakubczak and Tim Jones and Liyiming Ke and Devin LeBlanc and Sergey Levine and Adrian Li-Bell and Mohith Mothukuri and Suraj Nair and Karl Pertsch and Allen Z. Ren and Lucy Xiaoyang Shi and Laura Smith and Jost Tobias Springenberg and Kyle Stachowicz and James Tanner and Quan Vuong and Homer Walke and Anna Walling and Haohuan Wang and Lili Yu and Ury Zhilinsky},
      year={2025},
      journal={arXiv preprint arXiv:2504.16054}
}

@inproceedings{zhaoaloha,
  title={ALOHA Unleashed: A Simple Recipe for Robot Dexterity},
  author={Zhao, Tony Z and Tompson, Jonathan and Driess, Danny and Florence, Pete and Ghasemipour, Seyed Kamyar Seyed and Finn, Chelsea and Wahid, Ayzaan},
  booktitle={8th Annual Conference on Robot Learning},
  year={2024}
}

@inproceedings{liu2023libero,
  title={Libero: Benchmarking knowledge transfer for lifelong robot learning},
  author={Liu, Bo and Zhu, Yifeng and Gao, Chongkai and Feng, Yihao and Liu, Qiang and Zhu, Yuke and Stone, Peter},
  booktitle={Advances in Neural Information Processing Systems},
  pages={44776--44791},
  year={2023}
}

@inproceedings{chernyadev2025bigym,
  title={BiGym: A Demo-Driven Mobile Bi-Manual Manipulation Benchmark},
  author={Chernyadev, Nikita and Backshall, Nicholas and Ma, Xiao and Lu, Yunfan and Seo, Younggyo and James, Stephen},
  booktitle={Conference on Robot Learning},
  pages={4201--4217},
  year={2025}
}

@article{zhao2023learning,
  title={Learning fine-grained bimanual manipulation with low-cost hardware},
  author={Zhao, Tony Z and Kumar, Vikash and Levine, Sergey and Finn, Chelsea},
  journal={arXiv preprint arXiv:2304.13705},
  year={2023}
}

@article{chi2025diffusion,
  title={Diffusion policy: Visuomotor policy learning via action diffusion},
  author={Chi, Cheng and Xu, Zhenjia and Feng, Siyuan and Cousineau, Eric and Du, Yilun and Burchfiel, Benjamin and Tedrake, Russ and Song, Shuran},
  journal={The International Journal of Robotics Research},
  pages={1684--1704},
  year={2025},
}

@article{kim2024openvla,
  title={Openvla: An open-source vision-language-action model},
  author={Kim, Moo Jin and Pertsch, Karl and Karamcheti, Siddharth and Xiao, Ted and Balakrishna, Ashwin and Nair, Suraj and Rafailov, Rafael and Foster, Ethan and Lam, Grace and Sanketi, Pannag and others},
  journal={arXiv preprint arXiv:2406.09246},
  year={2024}
}

@inproceedings{zitkovich2023rt,
  title={Rt-2: Vision-language-action models transfer web knowledge to robotic control},
  author={Zitkovich, Brianna and Yu, Tianhe and Xu, Sichun and Xu, Peng and Xiao, Ted and Xia, Fei and Wu, Jialin and Wohlhart, Paul and Welker, Stefan and Wahid, Ayzaan and others},
  booktitle={Conference on Robot Learning},
  pages={2165--2183},
  year={2023},
}

@inproceedings{lin2024any,
  title={Any-step Dynamics Model Improves Future Predictions for Online and Offline Reinforcement Learning},
  author={Lin, Haoxin and Xu, Yu-Yan and Sun, Yihao and Zhang, Zhilong and Li, Yi-Chen and Jia, Chengxing and Ye, Junyin and Zhang, Jiaji and Yu, Yang},
  booktitle={The Thirteenth International Conference on Learning Representations},
  year={2024}
}

@article{gu2024review,
  title={A review of safe reinforcement learning: Methods, theories and applications},
  author={Gu, Shangding and Yang, Long and Du, Yali and Chen, Guang and Walter, Florian and Wang, Jun and Knoll, Alois},
  journal={IEEE Transactions on Pattern Analysis and Machine Intelligence},
  year={2024},
  publisher={IEEE}
}

@inproceedings{shi2023waypoint,
  title={Waypoint-Based Imitation Learning for Robotic Manipulation},
  author={Shi, Lucy Xiaoyang and Sharma, Archit and Zhao, Tony Z and Finn, Chelsea},
  booktitle={Conference on Robot Learning},
  pages={2195--2209},
  year={2023},
  organization={PMLR}
}

@book{lynch2017modern,
  title={Modern robotics},
  author={Lynch, Kevin M and Park, Frank C},
  year={2017},
  publisher={Cambridge University Press}
}

@inproceedings{kostrikovoffline,
  title={Offline Reinforcement Learning with Implicit Q-Learning},
  author={Kostrikov, Ilya and Nair, Ashvin and Levine, Sergey},
  booktitle={International Conference on Learning Representations},
  year={2021}
}

@inproceedings{lipmanflow,
  title={Flow Matching for Generative Modeling},
  author={Lipman, Yaron and Chen, Ricky TQ and Ben-Hamu, Heli and Nickel, Maximilian and Le, Matthew},
  booktitle={The Eleventh International Conference on Learning Representations},
  year={2022}
}

@article{zare2024survey,
  title={A survey of imitation learning: Algorithms, recent developments, and challenges},
  author={Zare, Maryam and Kebria, Parham M and Khosravi, Abbas and Nahavandi, Saeid},
  journal={IEEE Transactions on Cybernetics},
  year={2024},
  publisher={IEEE}
}

@article{ravichandar2020recent,
  title={Recent advances in robot learning from demonstration},
  author={Ravichandar, Harish and Polydoros, Athanasios S and Chernova, Sonia and Billard, Aude},
  journal={Annual review of control, robotics, and autonomous systems},
  pages={297--330},
  year={2020},
  publisher={Annual Reviews}
}

@inproceedings{zhao2023state,
  title={State-wise safe reinforcement learning: a survey},
  author={Zhao, Weiye and He, Tairan and Chen, Rui and Wei, Tianhao and Liu, Changliu},
  booktitle={Proceedings of the Thirty-Second International Joint Conference on Artificial Intelligence},
  pages={6814--6822},
  year={2023}
}

@inproceedings{shoemake1985animating,
  title={Animating rotation with quaternion curves},
  author={Shoemake, Ken},
  booktitle={Proceedings of the 12th annual conference on Computer graphics and interactive techniques},
  pages={245--254},
  year={1985}
}

@article{wang2025vla,
  title={Vla-adapter: An effective paradigm for tiny-scale vision-language-action model},
  author={Wang, Yihao and Ding, Pengxiang and Li, Lingxiao and Cui, Can and Ge, Zirui and Tong, Xinyang and Song, Wenxuan and Zhao, Han and Zhao, Wei and Hou, Pengxu and others},
  journal={arXiv preprint arXiv:2509.09372},
  year={2025}
}

@inproceedings{pham2019critically,
  title={Critically fast pick-and-place with suction cups},
  author={Pham, Hung and Pham, Quang-Cuong},
  booktitle={2019 International Conference on Robotics and Automation (ICRA)},
  pages={3045--3051},
  year={2019},
  organization={IEEE}
}

@article{kiyokawa2022challenges,
  title={Challenges for future robotic sorters of mixed industrial waste: a survey},
  author={Kiyokawa, Takuya and Takamatsu, Jun and Koyanaka, Shigeki},
  journal={IEEE Transactions on Automation Science and Engineering},
  pages={1023--1040},
  year={2022},
  publisher={IEEE}
}

@article{black2025real,
  title={Real-Time Execution of Action Chunking Flow Policies},
  author={Black, Kevin and Galliker, Manuel Y and Levine, Sergey},
  journal={arXiv preprint arXiv:2506.07339},
  year={2025}
}

@article{black2025training,
  title={Training-time action conditioning for efficient real-time chunking},
  author={Black, Kevin and Ren, Allen Z and Equi, Michael and Levine, Sergey},
  journal={arXiv preprint arXiv:2512.05964},
  year={2025}
}

@article{ma2025running,
  title={Running vlas at real-time speed},
  author={Ma, Yunchao and Zhou, Yizhuang and Yang, Yunhuan and Wang, Tiancai and Fan, Haoqiang},
  journal={arXiv preprint arXiv:2510.26742},
  year={2025}
}

@article{tang2025vlash,
  title={Vlash: Real-time vlas via future-state-aware asynchronous inference},
  author={Tang, Jiaming and Sun, Yufei and Zhao, Yilong and Yang, Shang and Lin, Yujun and Zhang, Zhuoyang and Hou, James and Lu, Yao and Liu, Zhijian and Han, Song},
  journal={arXiv preprint arXiv:2512.01031},
  year={2025}
}

@inproceedings{wu2025device,
  title={On-device diffusion transformer policy for efficient robot manipulation},
  author={Wu, Yiming and Wang, Huan and Chen, Zhenghao and Pang, Jianxin and Xu, Dong},
  booktitle={Proceedings of the IEEE/CVF International Conference on Computer Vision},
  pages={14073--14083},
  year={2025}
}

@article{yang2025efficientvla,
  title={EfficientVLA: Training-Free Acceleration and Compression for Vision-Language-Action Models},
  author={Yang, Yantai and Wang, Yuhao and Wen, Zichen and Zhongwei, Luo and Zou, Chang and Zhang, Zhipeng and Wen, Chuan and Zhang, Linfeng},
  journal={arXiv preprint arXiv:2506.10100},
  year={2025}
}

@article{gao2025compressor,
  title={Compressor-VLA: Instruction-Guided Visual Token Compression for Efficient Robotic Manipulation},
  author={Gao, Juntao and Ye, Feiyang and Zhang, Jing and Qian, Wenjing},
  journal={arXiv preprint arXiv:2511.18950},
  year={2025}
}

@article{spearman1961proof,
  title={The proof and measurement of association between two things.},
  author={Spearman, Charles},
  year={1961},
  publisher={Appleton-Century-Crofts}
}

@inproceedings{seo2023multi,
  title={Multi-view masked world models for visual robotic manipulation},
  author={Seo, Younggyo and Kim, Junsu and James, Stephen and Lee, Kimin and Shin, Jinwoo and Abbeel, Pieter},
  booktitle={International Conference on Machine Learning},
  pages={30613--30632},
  year={2023},
  organization={PMLR}
}
\bibliographystyle{icml2026}

%%%%%%%%%%%%%%%%%%%%%%%%%%%%%%%%%%%%%%%%%%%%%%%%%%%%%%%%%%%%%%%%%%%%%%%%%%%%%%%
%%%%%%%%%%%%%%%%%%%%%%%%%%%%%%%%%%%%%%%%%%%%%%%%%%%%%%%%%%%%%%%%%%%%%%%%%%%%%%%
% APPENDIX
%%%%%%%%%%%%%%%%%%%%%%%%%%%%%%%%%%%%%%%%%%%%%%%%%%%%%%%%%%%%%%%%%%%%%%%%%%%%%%%
%%%%%%%%%%%%%%%%%%%%%%%%%%%%%%%%%%%%%%%%%%%%%%%%%%%%%%%%%%%%%%%%%%%%%%%%%%%%%%%
\appendix
\onecolumn
\allowdisplaybreaks
\section{Proof for Proposition \ref{prop:performance}} \label{app:proof1}
% Proposition 1 (Performance Monotonicity). Let $\pi_{\text{sched}}$ be the policy induced by the scheduler which selects action $k$ at state $o_t$, and $\pi_{\text{base}}$ be the base policy which selects $A_t$. If the scheduler satisfies the zero-violation constraint such that $c_q(s_t, k_t) \le 0$ for all visited states, then the performance of the accelerated policy is guaranteed to be no worse than the base policy, i.e., $J(\pi_{\text{sched}}) \ge J(\pi_{\text{base}})$.
We aim to prove that: $\forall s,Q^{\pi}(s, A^k) \ge Q^{\pi}(s, A)$ (zero-violation), implies $V^{\pi'}(s_0) \ge V^{\pi}(s_0)$ (success rate guarantee). In the following derivation, we assume $\gamma=1$.

First, we must rigorously define what the ``Action Chunk'' $A$ and its Q-value represent in terms of atomic, low-level control actions. Let an action chunk $A$ consist of a sequence of atomic actions $u$ over a physical duration $L$.\begin{itemize}\item Original Chunk $A$: Sequence $\{u_1, u_2, \dots, u_L\}$.\item Accelerated Chunk $A^k$: Sequence $\{u'_1, u'_2, \dots, u'_{L'}\}$ where $L' = L/k$.\end{itemize}The execution of a chunk is not a single jump, but a trajectory of atomic state transitions. Let $s_{\tau, i}$ denote the state at the $i$-th atomic step within the execution of chunk $A_\tau$ (where $\tau$ is the chunk index).The Q-value of a chunk $A$ under policy $\pi$ is defined as the sum of atomic rewards within the chunk plus the value of the state after the chunk finishes:\begin{equation} \label{eq:atomic_expansion}Q^{\pi}(s_\tau, A) = \mathbb{E}\left[ \underbrace{\sum_{i=1}^{L} r(s_{\tau, i}, u_i)}_{\text{Intra-chunk Reward}} + V^{\pi}(s_{\tau+1}) \right],\end{equation}where $s_{\tau+1}$ is the state reached after executing the last atomic action $u_L$. Similarly, for the accelerated chunk $A^k$:\begin{equation}Q^{\pi}(s_\tau, A^k) = \mathbb{E}\left[ \sum_{j=1}^{L'} r(s'_{\tau, j}, u'_j) + V^{\pi}(s'_{\tau+1}) \right].\end{equation}

We now prove that $V^{\pi'}(s) \ge V^{\pi}(s)$ for all states, which directly implies the final required inequality. We use mathematical induction (or recursive expansion) over the sequence of chunks. Let $V^{\pi'}(s)$ be the value of following the scheduler policy $\pi'$ (which always selects $A^k$). By definition:\begin{equation}V^{\pi'}(s_0) = Q^{\pi'}(s_0, A_0^k) = \mathbb{E} \left[ \sum_{j=1}^{L'_0} r(s'_{0, j}, u'_j) + V^{\pi'}(s_{1}) \right].\end{equation}We now show that $V^\pi(s_0)\leq V^{\pi'}(s_0)$ with recursive expansion of the value function: 
\begin{align}
    V^{\pi}(s_0) 
    &= Q^{\pi}(s_0, A_0) \le Q^{\pi}(s_0, A_0^k) \\
    &= \mathbb{E}_{s_1\sim \pi'} \left[ \sum_{j=1}^{L'_0} r(s'_{0,j}, u'_{0,j}) + V^{\pi}(s_1) \right] \\ 
    &= \mathbb{E}_{s_1\sim \pi'} \left[ \sum_{j=1}^{L'_0} r(s'_{0,j}, u'_{0,j}) + Q^{\pi}(s_1, A_1) \right]   \\
    &\le \mathbb{E}_{s_1\sim \pi'} \left[ \sum_{j=1}^{L'_0} r(s'_{0,j}, u'_{0,j}) + Q^{\pi}(s_1, A_1^k) \right] \\
    &= \mathbb{E}_{s_1, s_2\sim \pi'} \left[ \sum_{j=1}^{L'_0} r(s'_{0,j}, u'_{0,j}) + \sum_{m=1}^{L'_1} r(s'_{1,m}, u'_{1,m}) + V^{\pi}(s_2) \right]   \\
    &\quad \vdots \quad   \\
    \label{eq:expansion_logic}
    &\le \mathbb{E}_{\tau \sim \pi'} \left[ \sum_{t=0}^{\infty} \sum_{j=1}^{L'_t} r(s'_{t,j}, u'_{t,j}) \right] = V^{\pi'}(s_0).  
\end{align} 
\textbf{Remarks on $\gamma=1$}: We emphasize that the undiscounted setting ($\gamma=1$) is both physically motivated and mathematically essential for our derivation. First, since our primary metric is the task success rate under sparse rewards, setting $\gamma=1$ ensures that the value function $V^\pi(s)$ directly represents the probability of success, i.e., $V^\pi(s) = \mathbb{P}(\text{Success} | \pi, s)$. Second, $\gamma=1$ is a necessary condition for the validity of the telescoping sum in Eq. \ref{eq:expansion_logic}. In a variable-duration setting where the scheduler accelerates execution, the physical arrival time at any future state $s_{t+1}$ differs from that of the base policy. If $\gamma < 1$, the discount factors associated with $s_{t+1}$ would not align between the two policies, preventing the intermediate terms in the performance difference expansion from canceling out. By assuming $\gamma=1$, the value of a state becomes invariant to $L$ and $L'$, allowing for a rigorous proof of global performance preservation despite temporal downsampling.

\section{Proof for Proposition \ref{prop:penalty}}
\label{app:proof2}
Let $Q^*(s, k)$ denote the optimal action-value function. The maximum possible value of a safe trajectory is bounded by $V_{\max} = \sum_{t=0}^\infty \gamma^t K_{\max} = \frac{K_{\max}}{1-\gamma}$.
Consider an arbitrary state $s$.
If an action $k_{v}$ violates the constraint (i.e., $h(s, k)=1$), its Q-value is bounded by:
\begin{equation}
Q^*(s, k_v) = -\Omega + \gamma \mathbb{E}[V^*(s')] \le -\Omega + \gamma V_{\max}.
\end{equation}
Conversely, since valid actions yield at least a reward of 1 (assuming $k \ge 1$), the Q-value of the optimal safe action $k_{safe}$ satisfies $Q^*(s, k_{safe}) \ge 1+\gamma$.
To ensure the optimal policy never selects a violation, we require $Q^*(s, k_{safe}) > Q^*(s, k_v)$. It suffices to show:
\begin{equation}
1+\gamma > -\Omega + \frac{\gamma K_{\max}}{1-\gamma} \implies \Omega > \frac{\gamma K_{\max}}{1-\gamma} - 1-\gamma.
\end{equation}
Thus, setting $\Omega > \frac{\gamma K_{\max}}{1-\gamma}$ (a strictly stronger condition) guarantees that any violating action has a lower value than any valid action, compelling the optimal policy to strictly satisfy the safety constraint.
\newcommand{\tb}[1]{\textbf{#1}}
\definecolor{HighLightColor}{rgb}{1.0, 0.92, 0.8}

\begin{table*}[t!]
    \centering
    \caption{Performance comparison of different methods in all Bigym tasks. Best success rates and shortest lengths are \textbf{bolded}. Method with best success rate per task is \colorbox{HighLightColor}{highlighted}.}
    \label{tab:bigym_all_result_3}
    \setlength{\tabcolsep}{4pt} % 稍微调整列间距以适应
    \small
    \resizebox{\linewidth}{!}{
    \begin{tabular}{clccccc}
        \toprule
        \multicolumn{2}{c}{\textbf{Method}} & {\footnotesize Sandwich Remove} & {\footnotesize Take Cups} & {\footnotesize Put Cups} & {\footnotesize Dishwasher Open Trays} & {\footnotesize Move Plate}\\
        \midrule
        \multirow{4}{*}{ACT} & -base & (0.45, 340.5) & (0.15, 288.3) & (0.28, 320.3)  & ( 1.0, 275.0) & \cg{( \tb{0.58}, 194.8)}\\
                             & -ds2  & (0.48, 186.3) & (0.13, 178.0) & (0.36, 175.9)  & ( 1.0, 169.0) & (0.50, 155.0)\\
                             & +\textit{DemoSpeedup}  & (0.56, 171.5) & (0.10, 183.9) & (0.33, 169.3)  & (1.0, 156.0) & (0.28, \tb{71.7})\\
                             & +\textit{SuP(Ours)}  & \cg{( \tb{0.64}, \tb{155.9})} & \cg{( \tb{0.20}, \tb{176.6})}  & \cg{( \tb{0.38}, \tb{156.0})}  & \cg{(1.0, \tb{131.0})} & (0.49, 167.2)\\
        \midrule
        \multirow{4}{*}{DP}  & -base & (0.40, 376.8) & (0.07, 284.6) & \cg{( \tb{0.28}, 307.7)} & (0.57, 351.2) & (0.33, 261.1)\\
                             & -ds2  & (0.40, 209.6) & (0.12, 218.0) & (0.22, 181.5) & (0.48, 310.2) & (0.36, 203.3)\\
                             & +\textit{DemoSpeedup}  & (0.35, 199.3) & \cg{( \tb{0.21}, 239.5)} & (0.25, \tb{143.9}) & (0.94, \textbf{108.8}) & (0.38, 178.8) \\
                             & +\textit{SuP(Ours)} & \cg{(\tb{0.42}, \tb{179.9})} & (0.19, \tb{203.0}) & (0.27, 200.8) & \cg{( \tb{0.97}, 267.4)} & \cg{( \tb{0.39}, \tb{164.8})}\\
        \bottomrule
        \multicolumn{2}{c}{\textbf{Method}} & {\footnotesize Saucepan to Hob} & {\footnotesize Flip Cutlery} & {\footnotesize Cupboards Close All} & {\footnotesize Sandwich Flip} & {\footnotesize Dishwasher Close Trays}\\
        \midrule
        \multirow{4}{*}{ACT} & -base & (0.78, 334.1) & (0.35, 243.6) & ( 1.0, 449.8)  & (0.20, 406.7) & ( 1.0, 200.3)\\
                             & -ds2  & (0.47, 248.7) & (0.22, 257.6) & ( 1.0, 234.0)  & (0.19, 239.4) & ( 1.0, 117.0)\\
                             & +\textit{DemoSpeedup}  & (0.78, \tb{154.5}) & \cg{( \tb{0.39}, \tb{90.8})} & \cg{( \tb{1.0}, \tb{202.1})}  & (0.18, \tb{156.8}) & ( 1.0, 95.0)\\
                             & +\textit{SuP(Ours)}  & \cg{( \tb{0.88}, 174.3)} & (0.27, 138.7)  & ( 1.0, 212.1)  & \cg{( \tb{0.22}, 194.5)} & \cg{( \tb{1.0}, \tb{86.0})}\\
        \midrule
        \multirow{4}{*}{DP}  & -base & \cg{( \tb{0.69}, 426.1)} & (0.06, 368.0) & (0.90, 544.0) & (0.06, 436.0) & (0.94, 210.4)\\
                             & -ds2  & (0.58, 285.0) & (0.12, 356.7) & (0.75, 270.3) & (0.03, \tb{157.3}) & (0.53, 146.4)\\
                             & +\textit{DemoSpeedup}  & (0.60, \tb{157.8}) & (0.10, \tb{145.9}) & (0.60, 231.0) & \cg{(0.11, 294.5)} & \cg{(\tb{0.96}, \tb{108.3})} \\
                             & +\textit{SuP(Ours)} & (0.66, 332.7) & \cg{( \tb{0.16}, 389.9)} & \cg{( \tb{0.91}, \tb{146.2})} & ( \tb{0.11}, 311.5) & (0.72, 175.4)\\
        \bottomrule
        \multicolumn{2}{c}{\textbf{Method}} & {\footnotesize Pick Box} & {\footnotesize Drawers Close All} & {\footnotesize Drawers Open All} & {\footnotesize Dishwasher Close} & {\footnotesize Wall Cupboard Open}\\
        \midrule
        \multirow{4}{*}{ACT} & -base & ( 0.25, 372.0) & ( 1.0, 100.0) & ( 1.0, 325.5)  & ( 1.0, 175.0) & ( 1.0, 148.8)\\
                             & -ds2  & (0.04, 182.0) & ( 1.0, 52.0) & (0.99, 190.0)  & ( 1.0, 90.9) & ( 1.0, 74.5)\\
                             & +\textit{DemoSpeedup}  & (0.02, 173.0) & ( 1.0, 54.0) & (0.99, 163.0)  & \cg{( \tb{1.0}, \tb{83.9})} & (0.92, 64.0)\\
                             & +\textit{SuP(Ours)}  & \cg{(0.25, \tb{317.9})} & \cg{( \tb{1.0}, \tb{40.0})}  & \cg{( \tb{1.0}, \tb{152.2})}  & (1.0, 84.0) & \cg{( \tb{1.0}, \tb{65.0})}\\
        \midrule
        \multirow{4}{*}{DP}  & -base & (0.0, -) & ( 0.66, 118.9) & \cg{( \tb{0.89}, 478.1)} & \cg{( \tb{0.99}, 178.9)} & (0.89, 260.0)\\
                             & -ds2  & (0.0, -) & (0.54, 64.9) & (0.12, 412.7) & (0.54, \tb{64.9}) & (0.88, \tb{131.0})\\
                             & +\textit{DemoSpeedup}  & (0.0, -) & (0.37, \tb{49.2}) & (0.44, \tb{193.2}) & (0.9, 158.4) & ( 0.91, 147.2)\\
                             & +\textit{SuP(Ours)} & (0.0, -) & \cg{(\tb{0.66}, 65.2)} & (0.30, 442.9) & (0.95, 147.3) & \cg{( \tb{0.91}, 146.2)}\\
        \bottomrule
        \multicolumn{2}{c}{\textbf{Method}} & {\footnotesize Store Box} & {\footnotesize Wall Cupboard Close} & {\footnotesize Dishwasher Open} & {\footnotesize Sandwich Toast} & {\footnotesize Flip Cup}\\
        \midrule
        \multirow{4}{*}{ACT} & -base & (0.51, 454.4) & ( 1.0, 100.0) & ( 1.0, 389.0)  & (0.08, 596.9) & \cg{( \tb{0.53}, 312.7)}\\
                             & -ds2  & (0.41, 229.2) & ( 1.0, 52.0) & ( 1.0, 377.0)  & (0.07, 261.9) & (0.32, 182.4)\\
                             & +\textit{DemoSpeedup}  & (0.33, \tb{222.2}) & \cg{( \tb{1.0}, \tb{46.0})} & ( 1.0, 163.3)  & (0.09, \tb{157.1}) & (0.30, \tb{149.8})\\
                             & +\textit{SuP(Ours)}  & \cg{( \tb{0.53}, 231.2)} & ( 1.0, 54.0)  & \cg{(1.0, \tb{157.4})} & \cg{( \tb{0.13}, 171.2)} & (0.46, 181.0)\\
        \midrule
        \multirow{4}{*}{DP}  & -base & (0.25, 456.0) & ( 1.0, 96.0) & (0.57, 354.1) & (0.04, 426.0) & (0.01, 312.0)\\
                             & -ds2  & (0.39, 328.6) & ( 1.0, 57.5) & (0.56, 281.3) & (0.03, \tb{178.7}) & (0.03, 552.0)\\
                             & +\textit{DemoSpeedup}  & (0.14, \tb{296.0}) & \cg{( \tb{1.0}, \tb{47.5})} & (0.38, \tb{113.5}) & (0.01, 180.0) & (0.04, \tb{150.3})\\
                             & +\textit{SuP(Ours)} & \cg{( \tb{0.39}, 336.6)} & ( 1.0, 50.6) & \cg{( \tb{0.57}, 267.4)} & \cg{( \tb{0.07}, 240.9)} & \cg{( \tb{0.05}, 164.5)}\\
        \bottomrule
    \end{tabular}
}
\end{table*}

\section{Whole Results of Bigym}
\label{sec:whole_bigym}
We report the (Success Rate, Episode Length) pairs for all 20 tasks across both ACT and DP architectures. As shown in Tab. \ref{tab:bigym_all_result_3}, SuP achieves the best balance between efficiency and success rate in most tasks, outperforming both static downsampling and DemoSpeedup baselines across a wide range of manipulation skills.

% 定义高亮颜色 (仿照图片中的浅橙色)

\section{Simulation Experiment Detail}
\label{app:sim_exp_detail}
\subsection{Bigym}
Here, we provide details of the BiGym tasks: we utilize a total of 20 tasks, all set in a kitchen scenario. The task descriptions (which can serve as language prompts if required) are listed below:
(1) Sandwich Remove: Take the sandwich out of the frying pan.
(2) Take Cups: Take two cups out from the closed wall cabinet and put them on the table.
(3) Put Cups: Pick up cups from the table and put them into the closed wall cabinet.
(4) Dishwasher Open Trays: Pull out the dishwasher’s trays with the door initially open.
(5) Move Plate: Move the plate between two draining racks.
(6) Saucepan to Hob: Take the saucepan from the closed cabinet and place it on the hob.
(7) Flip Cutlery: Take the cutlery from the static holder, flip it, and place it back into the holder.
(8) Cupboards Close All: Close all drawers and doors of the kitchen set.
(9) Sandwich Flip: Flip the sandwich in the frying pan using the spatula.
(10) Dishwasher Close Trays: Push the dishwasher’s trays back with the door initially open.
(11) Pick Box: Pick up a large box from the floor and place it on the counter.
(12) Drawers Close All: Close all sliding drawers of the kitchen cabinet.
(13) Drawers Open All: Open all sliding drawers of the kitchen cabinet.
(14) Dishwasher Close: Push back all trays and close the door of the dishwasher.
(15) Wall Cupboard Open: Open doors of the wall cabinet.
(16) Store Box: Move a large box from the counter to the shelf in the cabinet below.
(17) Wall Cupboard Close: Close doors of the wall cabinet.
(18) Dishwasher Open: Open the dishwasher door and pull out all trays.
(19) Sandwich Toast: Use the spatula to put the sandwich on the frying pan and toast it.
(20) Flip Cup: Flip the cup initially positioned upside down on the table to an upright position.

\textbf{1. Observation Space (State Space)} \\
BiGym's observation space is hybrid, combining visual inputs, proprioceptive data, and (for bi-manual mode) base state, which is defined as:
\[
O = \{I_{\text{head}}, I_{\text{left}}, I_{\text{right}}, s_{\text{proprio}}\}
\]
\begin{itemize}
    \item \textbf{Visual Observations}: RGB images ($I_{\text{head}}, I_{\text{left}}, I_{\text{right}}$) from three cameras (forehead, left wrist, right wrist), with a default resolution of $84 \times 84$.
    \item \textbf{Proprioceptive State ($s_{\text{proprio}}$)}: The state space adopts the Bi-manual mode, with the low-dimensional state ranging from 60 to 70 dimensions, including joint angles, joint velocities, and base states (where the leg control is configured in floating base mode), etc.
\end{itemize}

\vspace{5pt}
\textbf{2. Action Space} \\
The action space $A \in \mathbb{R}^{16}$ in Bigym can be formularized as three parts:
\begin{equation*}
    A = \{A_{\text{arms}} (\mathbb{R}^{10}), A_{\text{base}} (\mathbb{R}^{4}), A_{\text{grip}} (\mathbb{R}^{2})\},
\end{equation*}
where $\{A_{\text{arms}}$ controls the qpos of the robot arm, $A_{\text{base}}$ controls the floating base (i.e. legs) of the robot and $A_{\text{grip}}$ controls the left and right gripper of the robot arm.
\vspace{5pt}

\textbf{3. Training of $\pi_{\text{base}}$} \\
% Biygm中，我们follow Demospeedup开源仓库中对ACT和DP的实现方式[https://github.com/lingxiao-guo/DemoSpeedup/tree/main/robobase]，由于没有现成checkpoint release，所以我们按照源代码在20个环境中重跑了算法，并选取在评估中胜率最高的模型作为我们的base policy，ACT和DP的训练超参数如这两张表所示，我们发现Bigym中DP的性能表现不如ACT方法。我们也尝试在Bigym上基于$\pi_{0.5}$训练模型，但发现完成任务的成功率普遍弱于ACT和DP，这可能说明该模型不适合用于全身控制中。
In the Bigym environment, we followed the implementation of ACT and DP from the DemoSpeedup's open-source repository (specifically the robobase folder)\footnote{\url{https://github.com/lingxiao-guo/DemoSpeedup/tree/main/robobase}}. 
Due to the absence of pre-released checkpoints, we retrained the algorithms across 20 environments according to the original source code. 
We selected the models that achieved the highest win rates during evaluation as our base policies. 
The training hyperparameters for ACT and DP are detailed in Tab.~\ref{tab:act_dp_hyperparams}, respectively. 
Our experiments revealed that the performance of DP in Bigym was generally inferior to that of ACT. 
We also attempted to train a model based on $\pi_{0.5}$, but we found that the success rates were lower than those of both ACT and DP in a lot of tasks.
This suggests that the model may not be suitable for whole-body control tasks. Consequently, we did not attempt to accelerate the VLA base policy in Bigym.

\begin{table*}[t]
    \centering
    \caption{Hyperparameters for ACT and DP in Bigym.}
    \label{tab:act_dp_hyperparams}
    
    % --- 左侧: ACT Hyperparameters ---
    \begin{minipage}{0.48\textwidth}
        \centering
        \caption*{ACT Hyperparameters}
        \begin{tabular}{lc}
            \toprule
            Hyperparameter & ACT \\
            \midrule
            Learning Rate & 1e-5 \\
            Weight Decay & 1e-4 \\
            Batch Size & 64 \\
            Chunk Size ($k$) & 24 \\
            Feedforward Dim & 3200 \\
            Hidden Dim & 512 \\
            Encoder Layers & 4 \\
            Decoder Layers & 7 \\
            Attention Heads & 8 \\
            Dropout & 0.1 \\
            \bottomrule
        \end{tabular}
    \end{minipage}
    \hfill % 填充中间间距，使两个表格撑满一行
    % --- 右侧: DP Hyperparameters (No Aloha) ---
    \begin{minipage}{0.48\textwidth}
        \centering
        \caption*{DP Hyperparameters}
        \begin{tabular}{lc}
            \toprule
            Hyperparameter & DP \\
            \midrule
            Learning Rate & 1e-4 \\
            Weight Decay & 1e-6 \\
            Batch Size & 64 \\
            Observation Horizon & 2 \\
            Action Horizon & 24 \\
            Diffusion Steps & 100 \\
            Noise Scheduler & DDPM \\
            Kernel Size & 5 \\
            Vision Model & MVT \cite{seo2023multi} \\
            Down Dims & [256,512,1024] \\
            \bottomrule
        \end{tabular}
    \end{minipage}
\end{table*}

\subsection{Libero}

The Libero suite comprises four specialized sub-suites, each designed to isolate or integrate specific types of knowledge transfer for robot manipulation tasks, with distinct focuses and standardized language instruction patterns. Libero-spatial focuses on the transfer of declarative knowledge about spatial relationships, using instructions that specify spatial descriptors and target objects; Libero-object targets declarative knowledge about object concepts, with instructions centered on object names and containers; Libero-goal concentrates on procedural knowledge about task goals, featuring instructions that outline action-oriented tasks; Libero-long consists of long-horizon tasks involving entangled declarative and procedural knowledge transfer, with multi-step instructions that combine spatial, object, and goal concepts. Below is a detailed breakdown of the observation and action spaces common to or specific to each sub-suite, along with their core characteristics.

\textbf{1. Observation Space (State Space)} \\
Libero's observation space is hybrid, combining visual inputs, proprioceptive data, which is defined as:
\[
O = \{I_{\text{top}}, I_{\text{wrist}}, s_{\text{proprio}}\}
\]
\begin{itemize}
    \item \textbf{Visual Observations}: RGB images ($I_{\text{top}}, I_{\text{wrist}}$) from two cameras (top, wrist), with a default resolution of $224 \times 224$.
    \item \textbf{Proprioceptive State ($s_{\text{proprio}}$)}: The state space is an 8-dimensional low-dimensional joint state space.
\end{itemize}

\vspace{5pt}
\textbf{2. Action Space} \\
The action space is 7-dimensional, with 6-dimensional delta-EEF control, and 1-dimensional Gripper control.

\textbf{3. Detail of $\pi_{\text{base}}$} \\
% 在Libero中，我们使用已经开源好的Libero模型参数，对于$\pi_{0.5}$，遵循其开源仓库的instruction [https://github.com/Physical-Intelligence/openpi/tree/main/examples/libero] 就可以获取对应的模型参数。对于VLA-Adapter，遵循其开源仓库的instruction [https://github.com/OpenHelix-Team/VLA-Adapter] 就可以。$\pi_{0.5}$为4个task suite使用相同的模型权重，而VLA-Adapter则使用独立的权重。
In the Libero environment, we utilized the officially released pre-trained model checkpoints. 
For $\pi_{0.5}$, we obtained the corresponding model parameters by adhering to the instructions provided in its open-source repository\footnote{\url{https://github.com/Physical-Intelligence/openpi/tree/main/examples/libero}}. 
Similarly, for the VLA-Adapter, we followed the instructions outlined in its respective repository\footnote{\url{https://github.com/OpenHelix-Team/VLA-Adapter}}. 
Specifically, $\pi_{0.5}$ employs a shared set of model weights across all four task suites, whereas the VLA-Adapter utilizes independent weights for each suite. The training demonstration data is downloaded directly via HuggingFace\footnote{\url{https://huggingface.co/datasets/openvla/modified_libero_rlds}}.

\begin{figure}
    \centering
    \includegraphics[width=0.8\linewidth]{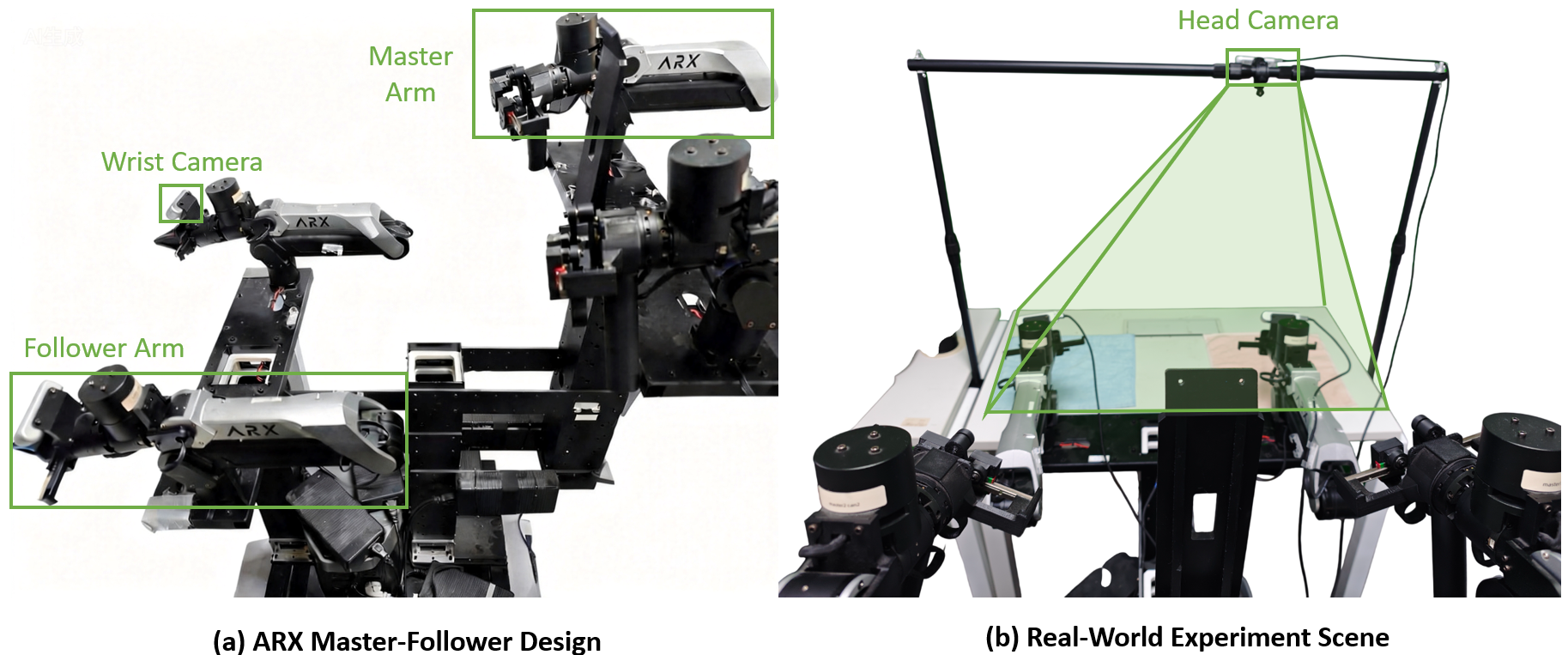}
    \caption{ARX5 illustration. (a) The master-follow design for data collection (b) The actual scene of our real-world experiment.}
    \label{fig:real_device}
\end{figure}

\section{Real-world Experiment Detail}
\label{app:real_exp}
\subsection{Hardware Setup}
The hardware configuration is detailed in Fig. \ref{fig:real_device}. We utilize the ARX5 robotic platform, a dual-arm system analogous to Aloha, consisting of two master arms and two puppet arms. Both arms were actively employed for dual-arm teleoperation and data collection. To provide visual feedback, a top-mounted RealSense D435i camera captures the RGB image observations required for the experiments.

\subsection{Details of Real-world Tasks}
\textbf{Fold Towel.}~~The scene consists of two towels of different colors or patterns placed on the tabletop. The robot is required to identify the target towel specified by a linguistic instruction and execute a folding sequence. This task tests the policy's ability to handle deformable objects and its grounding of language instructions in a multi-object scene.

\textbf{Arrange Table.}~~This task involves three plates and five objects initially distributed across them (arranged in a $2, 2, 1$ pattern). The robot must follow a three or four-step instruction to pick and place specific objects into designated plates. This task represents a long-horizon challenge requiring precise spatial reasoning and high-level planning.

\textbf{Stack Plates.}~~Three plates are placed separately on the table. The robot must stack them into a single pile following a specific order provided in the instruction (e.g., bottom-to-top sequence). This task emphasizes contact-rich manipulation and the strict maintenance of operational order.

\textbf{Training of $\pi_{\text{base}}$.}~~We collected a total of 200 high-quality demonstrations using teleoperation, with 50 trajectories in Fold Towel, 100 trajectories in Arrange Table and 50 trajectories in Stack Plates. We then utilized the $\pi_{0.5}$ model as the foundation. The model was fine-tuned on task-specific trajectoris to serve as $\pi_{\text{base}}$, ensuring reliable execution of the fundamental manipulation primitives.
\section{Details of SuP}
\label{app: detail of SuP}
In this section, we provide the detailed implementation of SuP, including how to downsample gripper action, how to calculate state deviation and the architecture of Recurrent World Model and scheduler. 
\subsection{Gripper Action Compensation for Downsampling}

Our methods rely on action chunk downsampling strategy that remain semantically aligned with the original one. While the downsampling strategy described in Sec. \ref{bg:ds} ensures that the robot's arm waypoints remain spatially consistent in the sense of they desired, gripper actions require separate consideration due to their binary nature and specific physical constraints. In most simulation environments, gripper actions are represented as binary signals (e.g., $<0$ for closed, $>0$ for open). Regardless of whether absolute or relative position control is used, standard downsampling causes a mismatch in the cumulative physical displacement of the gripper. For example, if a full grasp requires several consecutive closure commands, reducing the action frequency results in the gripper failing to reach the intended state in time, leading to failed grasps.To resolve this inconsistency and maintain the success rate after downsampling, we applied the following task-specific compensations:
\begin{itemize}
    \item BiGym: We followed the method described in DemoSpeedup \cite{guo2025demospeedup} by increasing the control gain of the gripper, ensuring it responds more aggressively to the reduced number of commands.
    \item Libero: We doubled the magnitude (velocity) of each gripper action command. 
\end{itemize}
For instance, if a single original action resulted in a 0.1 cm closure, the adjusted action for a downsampling factor of 2 ($N=2$) produces a 0.2 cm closure. Although this adjustment is specifically tailored for a downsampling rate of 2, we found it to be a highly effective heuristic for maintaining physical state consistency. The necessity of gripper action compensation is quantitatively validated in Tab. \ref{tab:gripper}. Without the fix, naive downsampling ($N=2$) leads to a significant performance degradation, with the average success rate dropping from 96.9\% to 84.2\%, particularly in the libero-spatial task where the gripper often fails to secure objects due to insufficient closure displacement. By applying our proposed compensation—adjusting the gripper's response magnitude—the ``-ds2'' variant recovers the average success rate to 92.6\% while maintaining a high inference speedup (1.72$\times$). This results in a much more robust balance between efficiency and task reliability.

\begin{table}[h]
    \centering
    \caption{Ablation study of gripper action compensation on Libero benchmarks. We compare the original policy ($pi_{0.5}$) with downsampled versions ($N=2$) before and after applying the gripper fix.}
    \label{tab:gripper}
    \begin{tabular}{lccccc}
        \toprule
        \textbf{Method} & \textbf{spatial} & \textbf{long} & \textbf{goal} & \textbf{object} & \textbf{Average} \\
        \midrule
        $\pi_{0.5}$ (Original) & 0.988, 105.3 & 0.924, 267.9 & 0.980, 113.1 & 0.982, 138.1 & 0.969, 1.00$\times$ \\
        \midrule
        No Grip Fix (-ds2) & 0.708, 77.0 & 0.818, 167.4 & 0.888, 66.2 & 0.954, 84.4 & 0.842, 1.58$\times$ \\
        With Grip Fix (-ds2) & 0.914, 67.9 & 0.874, 153.4 & 0.952, 67.6 & 0.970, 75.0 & 0.928, 1.72$\times$ \\
        \bottomrule
    \end{tabular}
\end{table}
\subsection{Calculation of State Deviation}
\label{app:state_deviation_calc}
In this section, we detail calculation of the state deviation metric $\mathcal{E}$, which serves as the core criterion for the switching logic within our Speedup Patch (SuP) framework. The calculation of state deviation relies on the formal representation of the robot's spatial configuration via the End-Effector (EEF) pose. The EEF pose is defined as a combination of its 3D Cartesian coordinates $(x, y, z)$ and its orientation, represented internally as a unit quaternion to avoid singularities. To evaluate the fidelity of the robot's motion during downsampled execution with a rate $k$, we determine the "expected" state at any intermediate sub-step $i \in \{1, \dots, k-1\}$ through pose interpolation between two consecutive reference waypoints $e_t$ and $e_{t+k}$ produced by the base policy. Specifically, the reference position is obtained via linear interpolation, while the reference orientation is computed using Normalized Linear Interpolation (NLERP) \cite{shoemake1985animating}. This approach ensures that the interpolated orientation remains on the unit hypersphere by normalizing the result of a linear interpolation between the two reference quaternions, providing a computationally efficient approximation of the shortest rotation path.

% 使用插值对齐降采样前后的EEF轨迹后，我们现在叙述如何计算两个EEF之间的距离$d(e_{curr},e_{ref})$. EEF距离由欧几里得位置和旋转距离组成。其中旋转距离定义为由有当前姿态到目标姿态的最小旋转距离，可以通过四元数计算得到。具体来说，两个EEF之间的距离定义为：
% $$d_{t+i} = \sqrt{(x_{curr} - x_{ref})^2 + (y_{curr} - y_{ref})^2 + (z_{curr} - z_{ref})^2} + \lambda \cdot \arccos(|\langle q_{curr}, q_{ref} \rangle|)$$
% 其中。。。
To evaluate the fidelity of the generated trajectories, we define a composite distance metric $d(e_{curr}, e_{ref})$ that measures the discrepancy between the current and reference end-effector (EEF) states. This distance comprises two components: the Euclidean distance for translational position and the geodesic distance for rotational orientation. The total distance at step $t+i$ is formulated as:$$d_{t+i} = \frac12\sqrt{(x_{curr} - x_{ref})^2 + (y_{curr} - y_{ref})^2 + (z_{curr} - z_{ref})^2} + \cdot \arccos(|\langle q_{curr}, q_{ref} \rangle|)$$where $\mathbf{p} = [x, y, z]^\top$ represents the Cartesian coordinates and $q$ denotes the orientation expressed as a unit quaternion. The rotational term calculates the minimum angular displacement between the two orientations, using the absolute value of the inner product $\langle q_{curr}, q_{ref} \rangle$ to account for the antipodal property of quaternions.

From an implementation perspective, these geometric operations—including NLERP and geodesic distance calculations—are natively and efficiently supported by the \textit{scipy.spatial.transform.Rotation module} in the SciPy library.

\subsection{Network architecture}
\begin{figure}[h]
    \centering
    \includegraphics[width=0.6\linewidth]{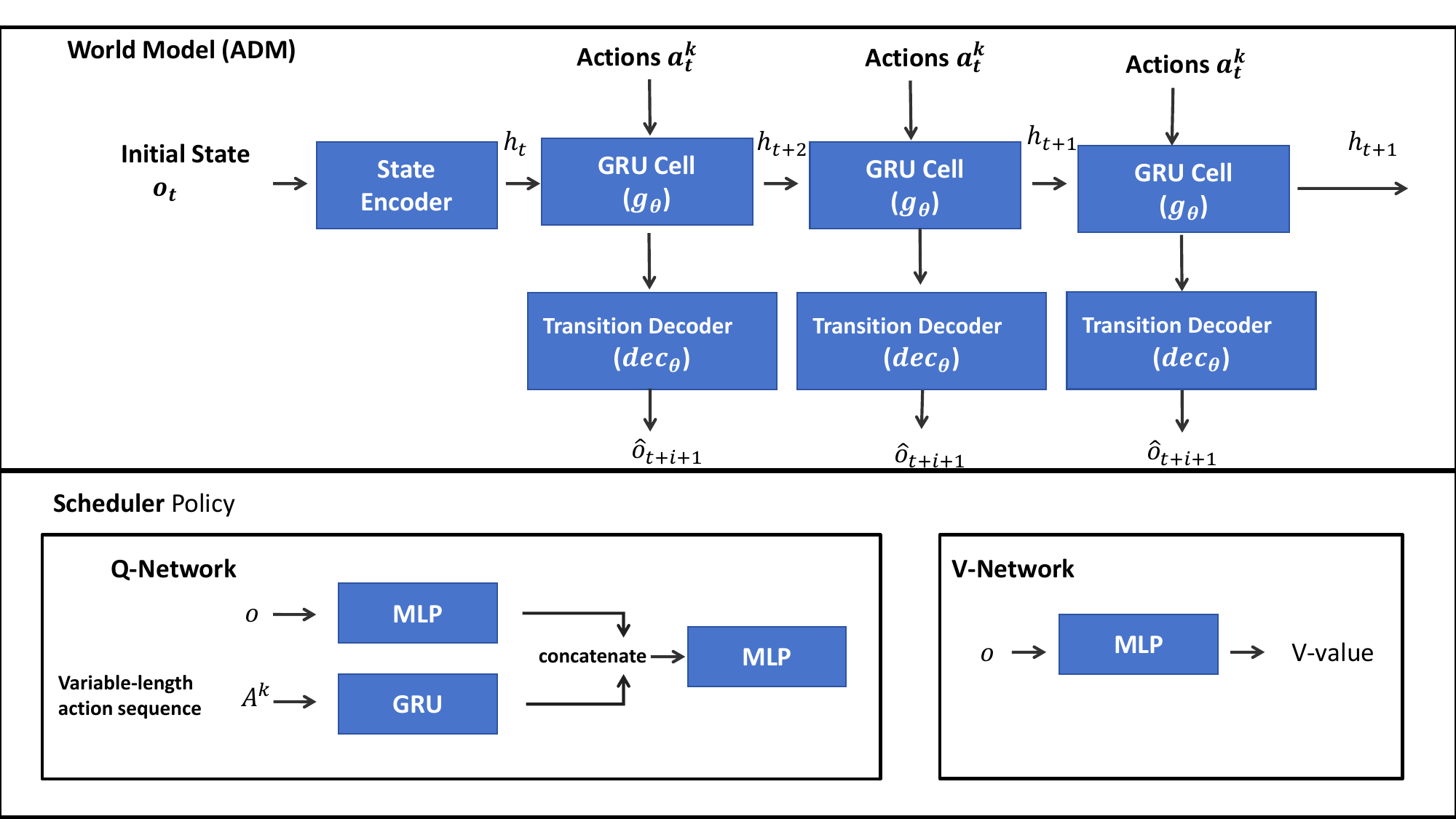}
    \caption{Network architecture of Recurrent World Model and Scheduler Policy.}
    \label{fig:architecture}
\end{figure}
\vspace{-1mm}

\textbf{Recurrent World Model.} The Any-Step Dynamics Model (ADM) is designed to predict future trajectories while bypassing the recursive error accumulation typical of auto-regressive transitions. The process begins by mapping the initial observation $o_t$ to a latent representation $h_t = \text{enc}_{\theta}(o_t)$ using a state encoder. This latent vector serves as the initial hidden state for a Gated Recurrent Unit (GRU), denoted as $g_{\theta}$. For each step $i \in \{0, \dots, L-1\}$, the GRU updates the hidden state via $h_{t+i+1} = g_{\theta}(h_{t+i}, a_{t+i}^k)$, conditioned on the previous latent state and the external action $a_{t+i}^k$. Crucially, a transition decoder $\text{dec}_{\theta}$ maps each latent state directly to a predicted observation $\hat{o}_{t+i+1}$. By decoupling the latent dynamics from the observation space—specifically by ensuring predicted observations are never fed back as inputs—the model maintains high trajectory fidelity and provides a stable foundation for counterfactual evaluation.

\textbf{Scheduler Policy.} The scheduler policy is implemented within the Implicit Q-Learning (IQL) framework, comprising separate Q and V networks. To handle the variable-length nature of the action sequences $A^k$, the Q-network employs a dual-stream architecture: a GRU processes the temporal dependencies of the action sequence, while a standard Multi-Layer Perceptron (MLP) encodes the current environment state $o$. The resulting features are concatenated and passed through a secondary MLP to produce the final Q-value. In contrast, the V-network utilizes a simplified architecture, consisting of a single MLP that maps the environment state $o$ directly to a state-value estimate. This design ensures the policy can effectively evaluate complex, multi-step action plans against the current environmental context.

\section{Hyperparameter of SuP}
% SuP
\begin{table}[h]
\centering
\label{tab:hyper}
\caption{Hyperparameter configurations of SuP in different experiment settings.}
\begin{tabular}{lccc}
\toprule
\textbf{Hyperparameter} & \textbf{BiGym} & \textbf{Libero} & \textbf{Real-world} \\ \midrule
Learning rate           & $3 \times 10^{-4}$    & $1 \times 10^{-4}$     & $1 \times 10^{-4}$         \\
Batch size              & 512            & 512             & 512                 \\
GRU hidden dimension    & 256            & 256             & 256                 \\
GRU layers              & 3              & 3               & 3                   \\
Chunk length            & 24             & 10              & 20                  \\
$k_{min}$               & 2              & 1               & 2                  \\
$k_{max}$               & 4              & 2               & 4                 \\
Epsilon ($\epsilon$)    & 0.01-0.02      & 0.01-0.02       & 0.02-0.04         \\
Expectile ($\tau$)      & 0.95           & 0.95            & 0.95                 \\
Penalty ($\Omega$)      & -5             & -2              & -1                  \\
Gamma ($\gamma$)        & 0.9            & 0.1             & 0.9                \\ \bottomrule
\end{tabular}
\end{table}

\section{Additional Visualization Results}
\label{app: Additional Visualization Results}
\begin{figure}[H]
    \centering
    \includegraphics[width=0.75\textwidth]{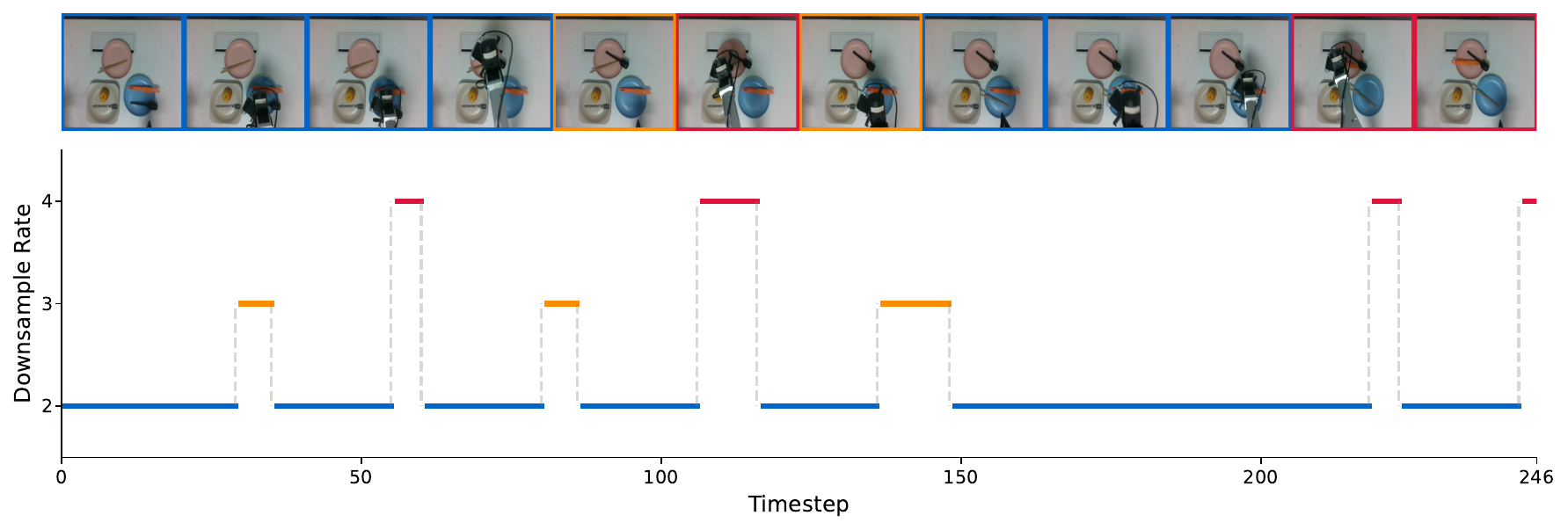}
    \caption{Arrange Table}  % 按需修改标题
    % \label{APP:}
\end{figure}

\begin{figure}[H]
    \centering
    \includegraphics[width=0.75\textwidth]{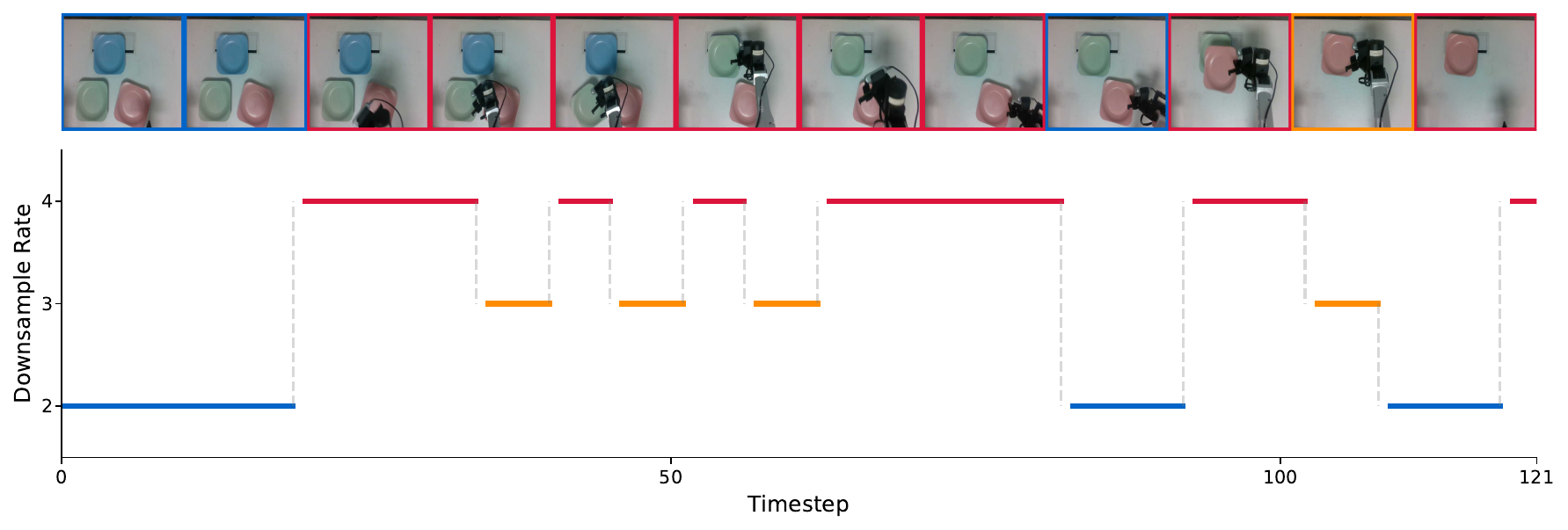}
    \caption{Stack Plate}  % 按需修改标题
    % \label{APP:}
\end{figure}
\begin{figure}[H]
    \centering
    \includegraphics[width=0.75\textwidth]{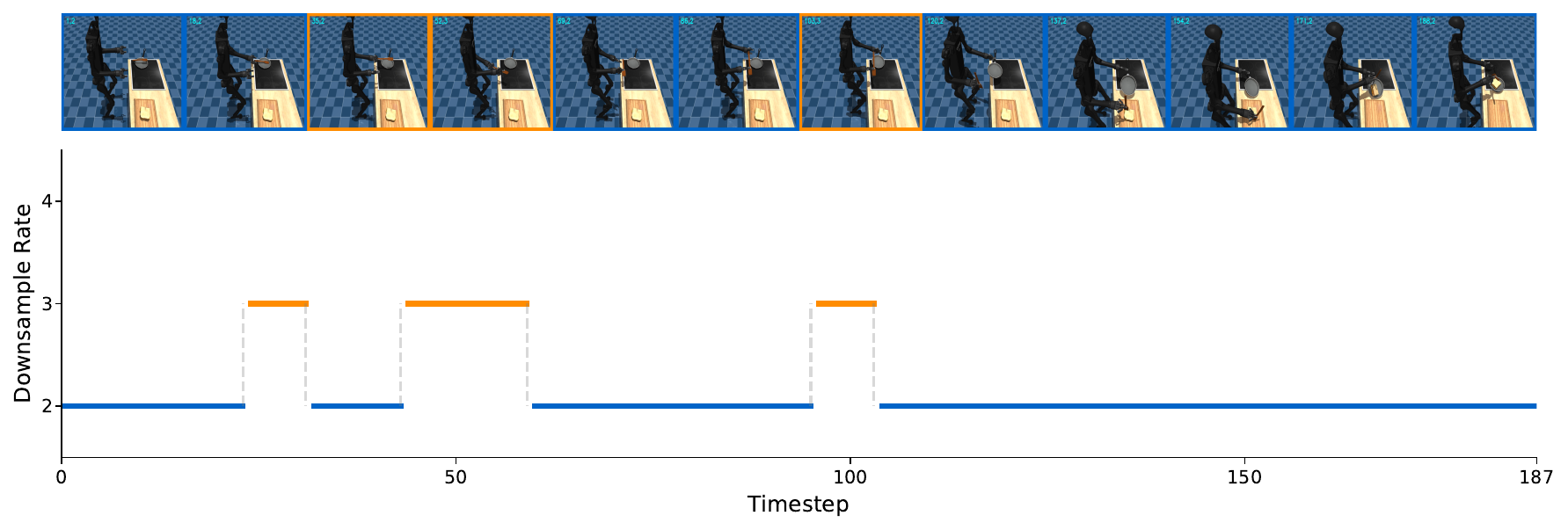}
    \caption{Sandwich Toast}  % 按需修改标题
    % \label{APP:}
\end{figure}
% \begin{figure}[H]
%     \centering
%     \includegraphics[width=0.75\textwidth]{figures/many_sup_samples/libero_long_cropped.pdf}
%     \caption{Libero Long}  % 按需修改标题
%     % \label{APP:}
% \end{figure}
\begin{figure}[H]
    \centering
    \includegraphics[width=0.75\textwidth]{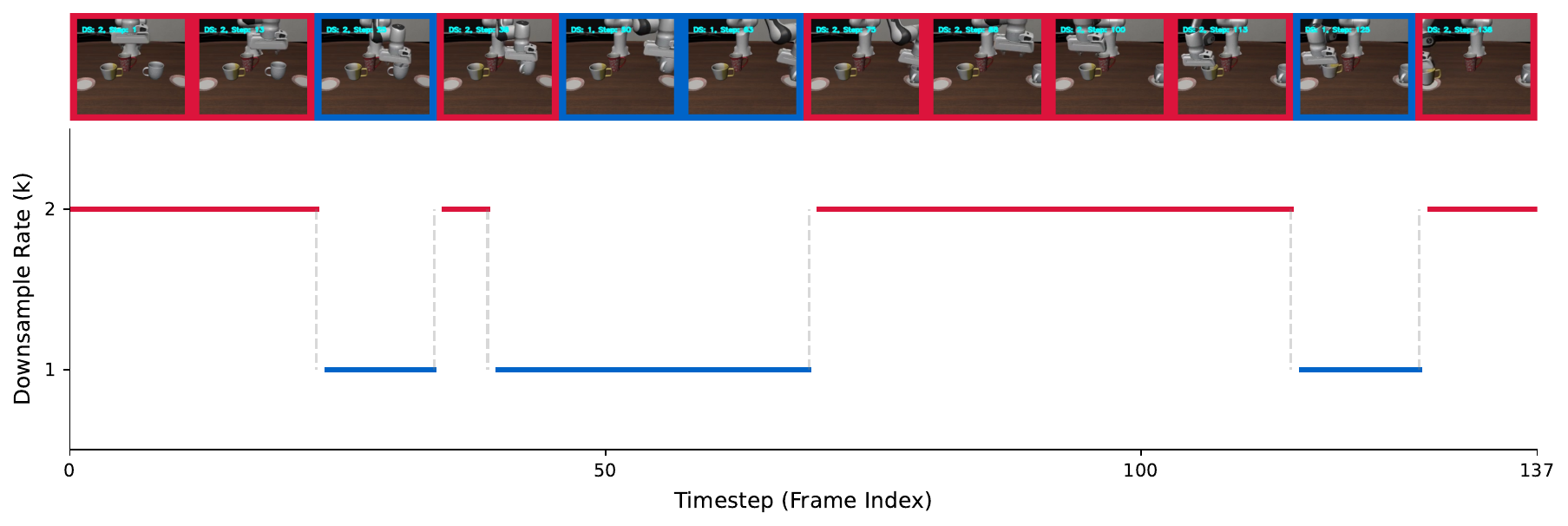}
    \caption{Libero Long}  % 按需修改标题
    % \label{APP:}
\end{figure}

\end{document}